\documentclass[runningheads]{llncs}
\usepackage{graphicx}
\usepackage{amsmath,amssymb}
\usepackage{color}
\usepackage{booktabs}
\usepackage{multirow}
\usepackage[maxbibnames=6,minbibnames=3,maxcitenames=2,mincitenames=1,style=numeric-comp,sorting=none]{biblatex}
\usepackage{pifont}
\usepackage{bm}
\usepackage[table]{xcolor}
\usepackage{colortbl}
\definecolor{bestcolor}{rgb}{0.85, 0.93, 0.85}
\definecolor{worstcolor}{rgb}{1.0, 0.90, 0.90}
\definecolor{lightgray}{gray}{0.93}
\definecolor{LightBlue}{HTML}{3498DB}
\definecolor{tamumaroon}{HTML}{500000}
\usepackage{tabularx}
\usepackage{threeparttable}
\usepackage{ragged2e}
\usepackage{makecell}
\newcolumntype{Y}{>{\raggedright\arraybackslash}X}
\newcolumntype{P}[1]{>{\raggedright\arraybackslash}p{#1}}
\newcolumntype{L}[1]{>{\RaggedRight\arraybackslash}p{#1}}

\usepackage{array}
\PassOptionsToPackage{hyphens}{url}
\usepackage[hidelinks,breaklinks=true]{hyperref}
\usepackage{caption}
\usepackage{tikz}
\usepackage{siunitx}
\usepackage{microtype}
\usepackage{enumitem}
\usepackage{fontawesome5}
\usepackage{float}
\usepackage{xstring}  
\AtBeginBibliography{\raggedright\small}
\newcommand{\ttt}[1]{\texttt{\hyphenchar\font=`\-\relax #1}}
\newcommand{\blueurl}[1]{%
  \StrSubstitute{#1}{https://}{}[\stripped]%
  \href{#1}{\textcolor{LightBlue}{\nolinkurl{\stripped}}}%
}
\makeatletter
\renewcommand{\@fnsymbol}[1]{\ifcase#1\or\hspace{0pt}\else\hspace{0pt}\fi}
\makeatother
\begin{document}
\pagestyle{headings}
\title{IRIS: A Real-World Benchmark for Inverse Recovery and Identification of Physical Dynamic Systems from Monocular Video}
\titlerunning{IRIS}
\authorrunning{Khanbayov, Barhdadi et al.}
\author{Rasul Khanbayov\inst{1}$^{*}$ \and Mohamed Rayan Barhdadi\inst{2}$^{*}$ \and \\ Erchin Serpedin\inst{2} \and Hasan Kurban\inst{1}\thanks{$^{*}$ Denotes equal contribution. Correspondence to: \texttt{hkurban@hbku.edu.qa}}.}
\institute{Hamad Bin Khalifa University \and Texas A\&M University}
\maketitle
\vspace{-10pt}
\begin{center}
\footnotesize{\faGlobe}\; Project Page:\\
\blueurl{https://kurbanintelligencelab.github.io/iris-bench/}

\end{center}
\vspace{-14pt}

\begin{abstract}
Unsupervised physical parameter estimation from video lacks a common benchmark: existing methods evaluate on non-overlapping synthetic data, the sole real-world dataset is restricted to single-body systems, and no established protocol addresses governing-equation identification. This work introduces IRIS, a high-fidelity benchmark comprising 240 real-world videos captured at 4K resolution and 60\,fps, spanning both single- and multi-body dynamics with independently measured ground-truth parameters and uncertainty estimates. Each dynamical system is recorded under controlled laboratory conditions and paired with its governing equations, enabling principled evaluation. A standardized evaluation protocol is defined encompassing parameter accuracy, identifiability, extrapolation, robustness, and governing-equation selection. Multiple baselines are evaluated, including a multi-step physics loss formulation and four complementary equation-identification strategies (VLM temporal reasoning, describe-then-classify prompting, CNN-based classification, and path-based labelling), establishing reference performance across all IRIS scenarios and exposing systematic failure modes that motivate future research. The dataset, annotations, evaluation toolkit, and all baseline implementations are publicly released.

\end{abstract}

\section{Introduction}
\label{sec:intro}

Extracting physical parameters from video observations constitutes a fundamental inverse problem in computational science, with applications spanning trajectory prediction, biological system characterization, and physical model validation~\cite{brunton2016discovering, hofherr2023neural, jaques2019physics, schmidt2009distilling}. Although recent unsupervised methods have demonstrated encouraging results in estimating parameters of known governing equations from video~\cite{hofherr2023neural, jaques2019physics}, progress is constrained by the absence of diverse, high-fidelity, real-world benchmark datasets capable of supporting rigorous evaluation and systematic comparison. Existing datasets suffer from several compounding limitations. The majority of prior work evaluates on purely synthetic data, where objects appear in controlled colors against black backgrounds~\cite{de2018end, jaques2019physics, velivckovic2022reasoning}. The sole notable real-world dataset, Delfys75~\cite{garcia2025learning}, provides 75 videos at 1920$\times$1080 resolution across five dynamical systems but covers only single-body dynamics, omitting the multi-body collision phenomena central to classical mechanics. No existing benchmark includes scenarios in which physical objects collide and transfer momentum, leaving a significant blind spot in evaluating method capabilities. \\

\begin{figure}[t]
\centering
\includegraphics[width=\linewidth]{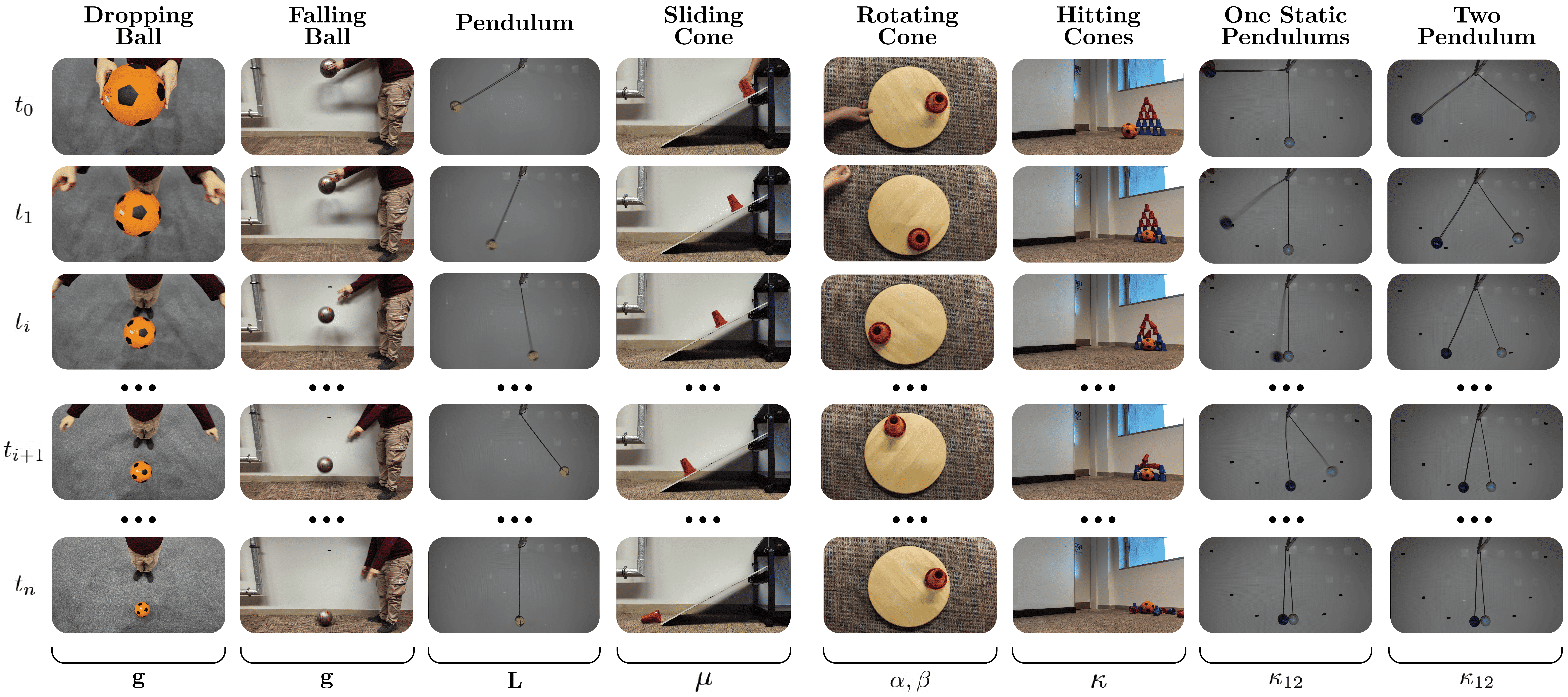}
\caption{\textbf{Overview of the IRIS benchmark.} Each column corresponds to one of the eight dynamical phenomena; rows show temporally ordered frames sampled from a representative video clip. \emph{Left single-body phenomena:} dropping ball, falling ball, pendulum, and sliding cone. \emph{Right phenomena unique to IRIS:} rotating cone, hitting cones, two pendulums, and one static pendulum. The bottom row indicates the physical parameters targeted for estimation. All videos are recorded at 4K resolution and 60\,fps under controlled laboratory conditions with independently measured ground-truth parameters.}
\label{fig:pipeline}
\end{figure}

These gaps are addressed by IRIS (Interaction and Real-world Inverse physics Sequences), a dataset designed as a rigorous benchmark for unsupervised physical parameter estimation from video. IRIS retains and extends the single-body scenarios of prior work (dropping ball, falling ball, sliding cone, and pendulum) while introducing three novel multi-body interaction scenarios: a ball striking a pyramid of cones (hitting\_\allowbreak{}cones), two pendulums released simultaneously that collide and dissipate (two\_\allowbreak{}moving\_\allowbreak{}pendulum), and a static pendulum struck by a moving one (two\_\allowbreak{}moving\_\allowbreak{}pendulum\_\allowbreak{}one\_\allowbreak{}static) and a new dynamic rotating cone on a wooden slide (rotating\_\allowbreak{}cone). These scenarios test recovery of momentum transfer, contact forces, and coupled dynamics, phenomena entirely absent from prior benchmarks. Beyond the dataset itself, a standardized evaluation protocol advances over the Delfys75 baseline along multiple dimensions. First, evaluation axes are formally defined covering parameter accuracy, governing-equation selection, identifiability, robustness, and extrapolation. Second, four complementary equation-identification strategies are benchmarked, providing the first systematic comparison of automatic equation routing for physics identification. Third, a multi-step physics loss formulation is evaluated as a baseline enhancement. Fourth, baseline evaluation on IRIS exposes a critical gradient-flow bug in the state-of-the-art latent-space pipeline~\cite{garcia2025learning}, demonstrating the diagnostic value of rigorous benchmarking. \\

\noindent In summary, our contributions are as follows: \\

\begin{enumerate}[leftmargin=*,nosep]
    \item \textbf{IRIS benchmark dataset:} A high resolution real-world dataset comprising eight dynamical systems, including novel single body motion and three multi-body interaction scenarios.
    \item \textbf{Standardized evaluation protocol:} A comprehensive framework with five formally defined axes (parameter accuracy, equation selection, identifiability, robustness, extrapolation).
    \item \textbf{Equation-identification benchmark:} The first systematic comparison of four equation-routing strategies: temporal-reasoning VLM prompting, two-stage describe-then-classify prompting, supervised CNN classification, and path-based oracle labelling.
    \item \textbf{Multi-step rollout loss and stability analysis:} A multi-step training objective for equation identification, with a stability analysis across single- and multi-body systems that reveals where such losses help and where they fail.
\end{enumerate}

\section{Related Work}
\label{sec:related}

\noindent\textbf{Physical Parameter Estimation from Videos:} Estimating physical parameters from video constitutes an inverse problem approached through both supervised and unsupervised paradigms. Supervised methods~\cite{asenov2019vid2param, de2018end, watters2017visual, wu2015galileo, wu2017learning, yang2022learning, zheng2018unsupervised} achieve strong performance at the cost of requiring labelled datasets that are expensive and frequently infeasible to collect~\cite{toth2019hamiltonian}. Unsupervised approaches embed known governing equations within learned latent representations, employing frame reconstruction as a proxy training signal. PAIG~\cite{jaques2019physics} integrates a variational autoencoder with a physics engine in the latent space, enabling future frame prediction via a spatial transformer. NIRPI~\cite{hofherr2023neural} employs a differentiable ODE solver operating at the pixel level, though it requires ground-truth object masks during training. Both rely on frame reconstruction, which constrains the model to roto-translational motion dynamics. A complementary line of research focuses on learning dynamics models with physics-informed inductive biases. Neural ODEs~\cite{chen2018neural} provide a general framework for continuous-time dynamics. Hamiltonian Neural Networks~\cite{greydanus2019hamiltonian} and Lagrangian Neural Networks~\cite{cranmer2020lagrangian} enforce energy conservation, while graph neural network-based simulators~\cite{sanchez2020learning, battaglia2016interaction} learn message-passing dynamics for multi-body systems. These operate on state-space inputs rather than raw pixels. Garcia et al.~\cite{garcia2025learning} replace frame reconstruction with a latent-space loss combining mean squared error and KL-divergence. However, their pipeline requires the governing equation family to be specified manually, and as identified through our baseline evaluation, contains a critical gradient-flow bug that prevents physics parameters from receiving gradient updates during training. \\

\noindent\textbf{Video Benchmarks for Physical Understanding:} The majority of existing methods are evaluated on synthetic datasets~\cite{de2018end, jaques2019physics, velivckovic2022reasoning, watters2017visual, yang2022learning, zheng2018unsupervised}. Physics101~\cite{wu2016physics} provides real-world recordings but lacks ground-truth parameters. VideoPhy~\cite{bansal2024videophy} and Physics-IQ~\cite{motamed2026generative} evaluate physical plausibility in generative models rather than quantitative parameter estimation. Delfys75~\cite{garcia2025learning} is the first real-world dataset for unsupervised parameter estimation with ground-truth annotations, but is restricted to single-body dynamics. Recent benchmarks including Morpheus~\cite{zhang2025morpheus}, RBench~\cite{guo2025rbench}, and WorldLens~\cite{liang2025worldlens} evaluate qualitative physical plausibility rather than quantitative parameter estimation. IRIS is complementary: it targets inverse recovery of continuous physical parameters with ground-truth supervision. \\

\noindent\textbf{Vision-Language Models for Physical Reasoning:} Large VLMs have demonstrated strong capabilities in visual scene understanding~\cite{roumeliotis2023chatgpt, liu2023visual, zhang2024llava}. Recent work has explored VLMs for physical reasoning: VideoPhy~\cite{bansal2024videophy} fine-tunes VLM-based evaluators for physical law adherence, Physics Context Builders~\cite{balazadeh2025physics} produce structured physical scene descriptions, and VLMs have been applied to hypothesize governing functional forms~\cite{liu2025mimicking}. Despite these advances, the application of VLMs to governing-equation identification for quantitative parameter estimation remains unexplored. IRIS benchmarks four complementary routing strategies for the first time. \\

\noindent\textbf{Governing Equation Discovery:} SINDy~\cite{brunton2016discovering} formulates equation discovery as sparse regression over candidate functions, with extensions to latent representations~\cite{champion2019data}. Symbolic regression methods~\cite{Schmidt2009DistillingFN} search expression spaces for closed-form laws, and multimodal LLMs have been applied to discover equations from video~\cite{li2025mllmbaseddiscoveryintrinsiccoordinates}. These methods require direct state measurements or high-quality latent representations. IRIS adopts a fixed ODE bank and benchmarks equation-family selection and parameter estimation as separable axes. \\

\noindent\textbf{Differentiable Contact and Rigid-Body Simulation:} Differentiable physics simulators that learn contact dynamics have advanced rapidly. ContactGaussian-WM~\cite{wang2026contact} combines 3D Gaussian splatting with a differentiable world model, while DiffSim~\cite{song2024diffsim} and related work~\cite{degrave2018differentiable, werling2021fastfeature} enable gradient-based optimization through differentiable simulation. These approaches model contact explicitly through complementarity or penalty-based formulations, in contrast to the simplified coupling coefficient $\kappa_{ij}$ employed in the IRIS ODE bank. Adapting such methods to monocular video and evaluating them on IRIS multi-body scenarios is a natural direction for future work. \\

\noindent\textbf{Latent ODE Stabilization:} The multi-step rollout instability observed on IRIS multi-body dynamics (Sec.~\ref{sec:exp_iris}) relates to a broader challenge in training latent ODEs over long horizons. Proposed stabilization techniques include path-length regularization~\cite{finlay2020trainneuralodeworld}, expressive latent priors~\cite{wang2021learningsolutionoperatorparametric}, and adaptive-step solvers with adjoint methods~\cite{chen2018neural}. These have not been evaluated for coupled multi-body ODE estimation from video; the catastrophic divergence exposed by IRIS (Sec.~\ref{sec:exp_iris}) suggests that such mechanisms may be essential for extending latent-space parameter estimation to interacting systems.

\section{The IRIS Benchmark}
\label{sec:dataset}

\subsection{Task Definition}
\label{sec:task}

IRIS targets \emph{physics identification from monocular video}: given frames $\mathcal{V} = \{I_1, \ldots, I_T\}$, the objectives are to (i)~identify the governing dynamical system from a bank of $K$ candidate ODEs $\{\mathcal{P}_1, \ldots, \mathcal{P}_K\}$ and (ii)~estimate its physical parameters $\boldsymbol{\gamma}$ without supervision or metadata. The observed motion is assumed driven by a low-dimensional latent state $\mathbf{z}_t \in \mathbb{R}^d$ evolving according to an ODE. For multi-body systems, the state extends to $\mathbf{Z}_t \in \mathbb{R}^{N \times d}$, where the number of interacting bodies $N$ is set \emph{a priori} from the experimental setup rather than estimated.

\subsection{Design Principles}
\label{sec:dataset_design}

IRIS is constructed around four principles: (1)~\textbf{controlled recording} under fixed lighting and stable camera placement; (2)~\textbf{high resolution and frame rate} at 4K/60\,fps; (3)~\textbf{independent ground truth} measured with calibrated instruments with uncertainty estimates; (4)~\textbf{repeated trials} with 10 recordings per setting, enabling variance analysis. The 4K resolution is a functional requirement, not an aesthetic one: a small ball of radius $r_0{=}0.04$\,m at 1.5\,m projects to ${\sim}3$ pixels at 1080p but ${\sim}12$ pixels at 4K, a $4\times$ pixel-density gain that reduces metric-scale calibration error and is a prerequisite for reliable falling-ball and pendulum-length recovery. The 10-trial design enables per-setting 95\% uncertainty bounds that are structurally impossible with fewer trials.

\subsection{Physical Phenomena and ODE Bank}
\label{sec:dataset_phenomena}

IRIS covers both \textbf{single-body} and \textbf{multi-body} phenomena. Table~\ref{tab:ode_bank} provides the explicit mapping between each phenomenon, its physical ODE, the ODE bank form used for estimation, and the target parameters. For the \emph{dropping ball}, the physical dynamics are governed by $z'' = -g$; however, the ODE bank employs a unified second-order linear form $z'' + \beta z' + \alpha z = 0$ following prior work~\cite{hofherr2023neural, jaques2019physics}, where $\alpha$ absorbs the gravitational constant in latent units and requires calibration to yield physical $g$. For multi-body phenomena, contact is modeled via a coupling coefficient $\kappa_{ij}$ that is continuously active, rather than through explicit impact mechanics; this simplification is discussed further below. \\

\begin{table}[t]
  \centering
  \caption{\textbf{Phenomenon--ODE mapping.} \textit{Physical ODE}: governing equation in SI units. \textit{ODE bank form}: parameterization used in the candidate library. \textit{Target params}: parameters estimated from video.}
  \label{tab:ode_bank}
  \footnotesize
  \setlength{\tabcolsep}{3.5pt}
  \renewcommand{\arraystretch}{1.25}
  \resizebox{\textwidth}{!}{%
  \begin{tabular}{@{}lllll@{}}
    \toprule
    \textbf{Phenomenon} & \textbf{Physical ODE} & \textbf{ODE bank form} & \textbf{Target params} & \textbf{Notes} \\
    \midrule
    \multicolumn{5}{@{}l}{\textit{Single-body}} \\[2pt]
    Dropping ball
      & $z'' = -g$
      & $z'' + \beta z' + \alpha z = 0$
      & $\alpha \!\approx\! g$,\,$\beta$
      & Local linearization \\
    Falling ball
      & $r(t) = r_0 f / (h_0 + \tfrac{1}{2}g t^2)$
      & 1\textsuperscript{st}-order apparent-size
      & $g$,\,$r_0$,\,$f$,\,$h_0$
      & Exact geometric proj. \\
    Sliding cone
      & $x'' = g(\sin\alpha - \mu\cos\alpha)$
      & Constant accel.\ $x'' = a$
      & $\alpha$\,[deg],\,$\mu$
      & Exact; $a$ via known $g$ \\
    Pendulum
      & $\theta'' + \zeta\theta' + (g/L)\sin\theta = 0$
      & Same (nonlinear $\sin\theta$)
      & $L$\,[m],\,$\zeta$,\,$g$
      & Exact; no small-angle \\
    Rotating cone
      & $\varphi'' + \beta\varphi' + \alpha\varphi = 0$
      & Same (damped torsional)
      & $\alpha$,\,$\beta$
      & Exact \\
    \midrule
    \multicolumn{5}{@{}l}{\textit{Multi-body}} \\[2pt]
    Hitting cones
      & Multi-body contact
      & $z_i'' + \zeta z_i' + \kappa \sum_j(z_i{-}z_j) = 0$
      & $\kappa$,\,$\zeta$
      & Simplified; see text \\
    Two mov.\ pend.
      & Eq.~\ref{eq:two_pendulums}
      & Same (coupled $\sin\theta$)
      & $L_i$,\,$\zeta_i$,\,$\kappa_{ij}$
      & Simplified; see text \\
    One stat.\ pend.
      & Eq.~\ref{eq:two_pendulums}
      & Same (coupled $\sin\theta$)
      & $L_i$,\,$\zeta_i$,\,$\kappa_{ij}$
      & Simplified; see text \\
    \bottomrule
  \end{tabular}%
  }
\end{table}

\textbf{Single-body dynamics} comprises five phenomena, each recorded under multiple experimental settings:
\begin{enumerate}
    \item \emph{Dropping ball}: a ball released from rest at height $h_0$\,[m], undergoing free fall. Three drop heights are recorded: 50\,cm, 100\,cm, and 150\,cm.
    \item \emph{Falling ball}: a ball filmed during free fall from above, with apparent radius governed by $r(t) = r_0 f / (h_0 + \tfrac{1}{2}g t^2)$. Three balls of different physical size are used (big: $r_0 = 0.11$\,m; mid: $r_0 = 0.07$\,m; small: $r_0 = 0.04$\,m).
    \item \emph{Sliding cone}: a cone sliding down an inclined plane, governed by $x''(t) = g(\sin\alpha - \mu\cos\alpha)$. Three inclination angles are recorded: $45^\circ$, $60^\circ$, and $80^\circ$.
    \item \emph{Pendulum}: a damped pendulum described by $\theta''(t) + \zeta\theta'(t) + (g/L)\sin\theta(t) = 0$. Three initial release angles are recorded: $20^\circ$, $45^\circ$, and $90^\circ$, with larger amplitudes probing the nonlinear $\sin\theta$ regime.
    \item \emph{Rotating cone}: a cone on a rotating wooden support, modeled as a damped torsional oscillator $\varphi''(t) + \beta\varphi'(t) + \alpha\varphi(t) = 0$. Three rotational speeds are defined by initial displacement: \emph{slow} (half-circle), \emph{mid} (one full circle), and \emph{fast} (two full circles).
\end{enumerate}

\textbf{Multi-body dynamics} introduces three novel interaction scenarios. \emph{Important modeling caveat:} contact between bodies is modeled via a continuously active coupling coefficient $\kappa_{ij}$ rather than through physically rigorous impact mechanics. This deliberate simplification ensures compatibility with existing latent-space estimation pipelines~\cite{hofherr2023neural, jaques2019physics, garcia2025learning}, which require smooth, differentiable dynamics. The recovered $\kappa_{ij}$ represents effective interaction strength rather than a physically grounded contact coefficient.

\begin{enumerate}
    \item \emph{Hitting cones}: a ball launched toward a 5-4-3-2-1 pyramid of 15 cones, testing recovery of contact-force and damping parameters from multi-object collision dynamics.
    \item \emph{Two moving pendulums}: two pendulums released simultaneously from the same initial angle $\theta_0$, colliding at the bottom. Three initial angles ($20^\circ$, $45^\circ$, $90^\circ$) vary the collision energy. The coupled system is governed by
    \begin{equation}
        \theta_i''(t) + \zeta_i\theta_i'(t) + \frac{g}{L_i}\sin\theta_i(t)
        + \kappa_{ij}(\theta_i(t) - \theta_j(t)) = 0,
        \quad i \neq j,
        \label{eq:two_pendulums}
    \end{equation}
    where $L_i$\,[m] is rope length, $\zeta_i$\,[s$^{-1}$] is damping, and $\kappa_{ij}$\,[s$^{-2}$] is the coupling coefficient.
    \item \emph{One static, one moving pendulum}: one pendulum hangs at rest while a second is released from the left, strikes the stationary pendulum, and both come to rest through repeated collisions. Three release angles ($20^\circ$, $45^\circ$, $90^\circ$) test sensitivity to initial energy and interaction asymmetry.
\end{enumerate}

Each single-body phenomenon contains 3 settings; multi-body phenomena contain 3 settings each, yielding 24 settings and 240 videos in total. Figure~\ref{fig:pipeline} presents representative frames. Table~\ref{tab:dataset_stats} summarizes the dataset.

\begin{table}[t]
  \centering
  \caption{\textbf{IRIS dataset summary.} Each setting contains 10 repeated recordings across diverse physical phenomena.}
  \label{tab:dataset_stats}
  \footnotesize
  \setlength{\tabcolsep}{4pt}
  \renewcommand{\arraystretch}{1.15}
  \begin{tabular}{@{}llcccc@{}}
    \toprule
    \textbf{Type} & \textbf{Phenomenon} & \textbf{\#\,Set.} & \textbf{\#\,Vid.} & \textbf{Dur.\,(s)} & \textbf{GT Params} \\
    \midrule
    \multirow{5}{*}{\textit{Single}}
      & Dropping ball    & 3 & 30  & 5   & $g,\,\beta$ \\
      & Falling ball     & 3 & 30  & 8   & $g,\,r_0,\,f,\,h_0$ \\
      & Sliding cone     & 3 & 30  & 5   & $\mu,\,\alpha$ \\
      & Pendulum         & 3 & 30  & 150 & $L,\,\zeta,\,g$ \\
      & Rotating cone    & 3 & 30  & 8   & $\alpha,\,\beta$ \\
    \midrule
    \multirow{3}{*}{\textit{Multi}}
      & Hitting cones    & 3 & 30  & 5   & $\kappa,\,\zeta$ \\
      & Two moving pend. & 3 & 30  & 6   & $L_i,\,\zeta_i,\,\kappa_{ij}$ \\
      & One static pend. & 3 & 30  & 20  & $L_i,\,\zeta_i,\,\kappa_{ij}$ \\
    \midrule
    \multicolumn{2}{@{}l}{\textbf{Total}}
      & \textbf{24} & \textbf{240} & -- & -- \\
    \bottomrule
  \end{tabular}
\end{table}

\subsection{Annotations and Ground Truth}
\label{sec:dataset_annotations}

For each recording IRIS provides: (1)~video clips in tensor format $(N, n_f, C, H, W)$; (2)~ground-truth physical parameters with uncertainty bounds in \textit{parameters.json}; (3)~equation family labels. Per-frame segmentation masks via SAM2~\cite{ravi2024sam2segmentimages} are provided for multi-object setups. A calibration checkerboard enables metric-scale recovery. \\

\textbf{Ground-truth measurement of damping and friction.}
\label{sec:gt_damping_friction}
Damping and friction coefficients carry larger uncertainty than geometric quantities. For the \emph{pendulum damping coefficient} $\zeta$, ground truth is obtained by fitting an exponential decay envelope $A(t) = A_0 e^{-\zeta t / 2}$ to peak amplitudes from extended-duration recordings (${\sim}150$\,s per clip), with uncertainty quantified as the 95\% confidence interval across the 10 trials per setting. For the \emph{friction coefficient} $\mu$, ground truth is obtained from $\mu = \tan\alpha - a_{\mathrm{measured}} / (g \cos\alpha)$, with uncertainty propagated from the standard deviation of $a_{\mathrm{measured}}$ across trials. For the \emph{torsional damping} $\beta$, the same exponential-envelope fitting is applied. Damping ground truth is derived from trajectory fits rather than independent physical measurement; this distinction is documented in \textit{parameters.json} via a \textit{measurement\_type} field (\textit{"direct"} vs.\ \textit{"fitted"}).

\newcommand{\cmark}{\textcolor{green!60!black}{\ding{51}}}
\newcommand{\xmark}{\textcolor{red!70!black}{\ding{55}}}

\begin{table}[t]
  \centering
  \caption{\textbf{Comparison with existing benchmarks} for physical parameter estimation from video.}
  \label{tab:benchmark_comparison}
  \footnotesize
  \setlength{\tabcolsep}{4pt}
  \renewcommand{\arraystretch}{1.15}
  \begin{tabular}{@{}lccccccc@{}}
    \toprule
    \textbf{Benchmark} & \textbf{Real} & \textbf{Resolution} & \textbf{Videos} & \textbf{Phen.} & \textbf{Multi} & \textbf{GT} & \textbf{Trials} \\
    \midrule
    PAIG synth.~\cite{jaques2019physics}  & \xmark & $50{\times}50$     & var. & 1  & \xmark & \xmark & --\, \\
    NIRPI synth.~\cite{hofherr2023neural} & \xmark & $100{\times}100$   & var. & 2  & \xmark & \xmark & --\, \\[2pt]
    Physics101~\cite{wu2016physics}       & \cmark & $1280{\times}720$  & 101  & 17 & \xmark & \xmark & 1   \\
    VideoPhy~\cite{bansal2024videophy}    & \cmark & var.               & 688  & -- & \xmark & \xmark & --\, \\
    Delfys75~\cite{garcia2025learning}    & \cmark & $1920{\times}1080$ & 75   & 5  & \xmark & \cmark & 5   \\[2pt]
    \textbf{IRIS (ours)}                  & \cmark & $3840{\times}2160$ & 240  & 8  & \cmark & \cmark & 10  \\
    \bottomrule
  \end{tabular}
\end{table}

\section{Evaluation Protocol}
\label{sec:eval_protocol}

A central contribution of IRIS is a method-agnostic evaluation protocol enabling fair comparison. For every video clip, independently: (i)~the equation family is determined; (ii)~the physics model is instantiated; (iii)~the model is trained on that clip from scratch. No parameters are shared across clips. The random seed is fixed for reproducibility. Five axes are defined. \emph{Parameter accuracy} (primary) measures closeness to ground truth via MAE:
$\mathrm{MAE} = n^{-1} \sum_{i=1}^{n} |\hat{\theta}_i - \theta^*|$,
with estimation standard deviation $\sigma_{\hat{\theta}}$ for stability. \emph{Equation selection accuracy} measures correct ODE assignment via overall and per-class accuracy with confusion matrices. \emph{Identifiability} is assessed through gradient norms $G_{\boldsymbol{\gamma}}^{(e)} = \|\nabla_{\boldsymbol{\gamma}}\mathcal{L}^{(e)}\|_2$ at epoch $e$ and the ODE residual $\mathcal{R} = (T{-}1)^{-1}\sum_t \|\hat{\mathbf{z}}_{t+1} - \mathcal{P}(\hat{\mathbf{z}}_t; \hat{\boldsymbol{\gamma}}, \Delta t)\|^2$. \emph{Robustness} measures consistency across design choices. \emph{Extrapolation} evaluates prediction beyond the training horizon via:
\begin{equation}
    \mathcal{E}_k = \|\hat{\mathbf{z}}_k - \tilde{\mathbf{z}}_k\|^2, \quad k > T_{\mathrm{train}},
    \label{eq:extrapolation}
\end{equation}
where $\hat{\mathbf{z}}_k$ is obtained by unrolling the learned ODE from the last training frame and $\tilde{\mathbf{z}}_k$ is the encoder output for the held-out frame $I_k$. 

All reported numbers are produced by a single evaluation script operating on fixed CSV outputs. Each table entry is traceable to a specific (phenomenon, setting, clip, seed) tuple. MAE values are computed over the test partition (2 clips per setting) unless stated otherwise. The evaluation code, CSV outputs, and random seeds are released with the dataset.

\section{Baselines}
\label{sec:baselines}

Representative baselines are evaluated to establish reference performance and expose failure modes. All follow the train-per-clip protocol with identical ground-truth sources.

\subsection{Latent-Space Baseline}
\label{sec:baseline_latent}

The primary baseline is the latent-space pipeline of Garcia et al.~\cite{garcia2025learning}: an encoder maps frames to latent states, a physics block implements a discrete-time ODE step, and training minimizes $\mathcal{L}_{\mathrm{1\text{-}step}} = M^{-1}\sum_i \|\hat{\mathbf{z}}_i - \tilde{\mathbf{z}}_i\|^2 + \mathcal{L}_{\mathrm{KL}}$, with path-based equation selection. \\

\textbf{Gradient-flow diagnosis.} Baseline evaluation on IRIS revealed a critical bug in the original Euler integrator: the position update omits the acceleration term, severing gradient flow to $\boldsymbol{\gamma}$. A \emph{corrected} variant restores the standard update $z_{t+1} = z_t + \Delta t\, z_t^{(1)} + \Delta t^2\, z_t^{(2)}$.

\subsection{Multi-Step Loss Variant}
\label{sec:baseline_multistep}

The corrected baseline is evaluated with a multi-step rollout loss enforcing agreement over $K$ future horizons:
\begin{equation}
    \mathcal{L}_{\mathrm{total}} =
    \sum_{k=1}^{K} w_k \left\|
    \hat{\mathbf{z}}_{t+k} - \tilde{\mathbf{z}}_{t+k}
    \right\|^2
    +\; \mathcal{L}_{\mathrm{KL}}(\hat{\mathbf{z}}),
    \label{eq:loss_multistep}
\end{equation}
where $\tilde{\mathbf{z}}_{t+k}$ is obtained by unrolling $k$ physics steps from $\hat{\mathbf{z}}_t$ and $w_k$ decays geometrically ($w = [1, 1, 0.5, 0.5, 0.25]$ for $K{=}5$). The rationale is that one-step supervision can underconstrain $\boldsymbol{\gamma}$: multi-step rollouts accumulate the effect of physical parameters over time, providing a denser gradient signal.

\subsection{Equation-Identification Strategies}
\label{sec:baseline_eq_id}

IRIS benchmarks five complementary strategies for governing-equation identification: \\

\textbf{(1) VLM temporal reasoning.} A VLM acts as a routing function $k^* = \mathrm{VLM}(\{I_{s_1}, \ldots, I_{s_m}\}; \mathcal{C}_1, \ldots, \mathcal{C}_K)$, where $m{=}5$ frames are sampled at $t = 0\%, 25\%, 50\%, 75\%, 100\%$ and $\mathcal{C}_k$ is a natural-language class description. Three VLM backends are evaluated: GPT-4V, LLaVA-Video-7B, and InternVL2-8B.\\

\textbf{(2) Describe-then-classify.} A two-stage VLM pipeline: the first call generates a free-form description of the observed motion; the second classifies the description into one of the $K$ equation families. This decouples visual perception from physical classification. \\

\textbf{(3) CNN video classifier.} A lightweight ResNet-18 (ImageNet-pretrained) with temporal mean pooling. Input is 5 frames sampled at evenly spaced temporal positions ($0\%, 25\%, 50\%, 75\%, 100\%$), each resized to $224{\times}224$; frame-level features are mean-pooled into a clip descriptor before an $n$-way linear classifier ($n{=}6$ on Delfys75, $n{=}8$ on IRIS). Labels are derived from folder structure, identical to the ground truth used for VLM evaluation. The model is trained with a random 80/20 split (seed~42) and we report best validation accuracy on the held-out 20\%. It represents the upper bound of in-distribution performance but requires retraining whenever the phenomenon set changes, whereas VLM strategies operate zero-shot. \\

\textbf{(4) Path-based labels (oracle).} Ground-truth equation family provided by folder structure, serving as the oracle condition. \\

\textbf{(5) ODE bank.} The library comprises: second-order linear (damped oscillator), first-order exponential decay, first-order nonlinear (Torricelli), constant acceleration (sliding), and coupled oscillator (Eq.~\ref{eq:two_pendulums}).


\noindent
\textbf{Limitation.} These calibration heuristics are phenomenon-specific and depend on the encoder's latent geometry, which may introduce systematic bias. IRIS provides calibration infrastructure (checkerboard frames, known object dimensions) but does not eliminate the underlying ambiguity.

\section{Experiments}
\label{sec:experiments}

Baselines are evaluated along five axes: parameter estimation accuracy (primary), governing-equation identification, identifiability, extrapolation, and component ablations. All experiments follow the train-per-clip protocol with a fixed random seed (42) and shared CSV output format.

\subsection{Experimental Setup}
\label{sec:exp_setup}

Three configurations are compared: (1)~\textbf{Baseline}: path-based equation selection, one-step loss, original uncorrected Euler integrator~\cite{garcia2025learning}; (2)~\textbf{+Corrected}: gradient-corrected Euler step, one-step loss; (3)~\textbf{+Multi-step}: corrected integrator with multi-step rollout loss ($K{=}5$). Additionally, four equation-identification strategies are evaluated independently.

\subsection{Physical Parameter Estimation on Delfys75}
\label{sec:exp_delfys}

The Delfys75 results below establish that the corrected baseline is meaningfully better than the original \emph{before} evaluating on IRIS, justifying our use of \textbf{+Corrected} (rather than the original~\cite{garcia2025learning}) as the IRIS reference baseline. Effect of Gradient Correction, Table~\ref{tab:delfys75_baseline_unified} reports MAE for the original and corrected baselines on Delfys75~\cite{garcia2025learning}. The most prominent result concerns the pendulum: the original baseline yields MAE of $95.07$\,m on \textit{pendulum\_45} for string length, compared to $3.80$\,m for the corrected variant ($\Delta\mathrm{MAE} = -91.27$\,m), a direct consequence of the severed gradient flow. Both variants report $\mathrm{MAE} \approx 0$ for $g$ on \textit{dropped\_ball/large} and \textit{free\_fall/mousepad} because the uncorrected integrator leaves $g$ at its initialization value $g_0 = 9.81$, which coincides with ground truth (see Table~\ref{tab:delfys75_baseline_unified} footnote). Substantial improvements are also observed on Torricelli drainage ($-3.16$, $-1.99$) and LED 2\,s decay ($-1.90$).
Effect of Multi-Step Loss on Delfys75, Table~\ref{tab:multistep_ablation} compares one-step and multi-step losses. Multi-step consistently reduces MAE for $g$ across free fall and dropped ball (largest: $\Delta\mathrm{MAE} = -0.90$ on \textit{free\_fall/table}). The \textit{led\_10s} setting degrades ($+5.37$) because slow dynamics cause rollout divergence.

\begin{table}[t]
  \centering
  \caption{\textbf{Parameter MAE on Delfys75:} original baseline (uncorrected Euler) vs.\ corrected variant. \colorbox{bestcolor}{\textbf{Bold}}~=~lower MAE. \emph{Takeaway:} the gradient correction is decisive, it cuts pendulum-length error from $95.07$ to $3.80$\,m.}
  \label{tab:delfys75_baseline_unified}
  \scriptsize
  \setlength{\tabcolsep}{3pt}
  \renewcommand{\arraystretch}{1.05}
  \begin{tabular}{@{}lllrrrr@{}}
    \toprule
    \textbf{Dynamics} & \textbf{Setting} & \textbf{Param} & \textbf{GT} &
    \textbf{Base.} & \textbf{Corr.} & $\bm{\Delta}$ \\
    \midrule
    Dropped ball & large & $g$\,[m/s\textsuperscript{2}] & 9.81 & 0.00\textsuperscript{\dag} & \cellcolor{bestcolor}\textbf{0.00} & $0.00$ \\
    Free fall & mousepad & $g$\,[m/s\textsuperscript{2}] & 9.81 & 0.00\textsuperscript{\dag} & \cellcolor{bestcolor}\textbf{0.00} & $0.00$ \\
    LED & led\_2s & $\gamma$ & 2.30 & 2.30 & \cellcolor{bestcolor}\textbf{0.40} & $-1.90$ \\
    \multirow{3}{*}{Pendulum}
        & pend.\_45 & $L$\,[m] & 0.45 & 95.07 & \cellcolor{bestcolor}\textbf{3.80} & $-91.27$ \\
        & pend.\_90 & $L$\,[m] & 0.90 & 50.15 & \cellcolor{bestcolor}\textbf{2.93} & $-47.22$ \\
        & pend.\_150 & $L$\,[m] & 1.50 & 29.83 & \cellcolor{bestcolor}\textbf{0.71} & $-29.12$ \\
    Sliding bl. & mid & $\mu$ & 0.21 & \cellcolor{bestcolor}\textbf{0.00} & \cellcolor{bestcolor}\textbf{0.00} & $0.00$ \\
    \multirow{2}{*}{Torricelli}
        & large & $k$ & 0.016 & 6.07 & \cellcolor{bestcolor}\textbf{2.91} & $-3.16$ \\
        & small & $k$ & 0.010 & 7.67 & \cellcolor{bestcolor}\textbf{5.69} & $-1.99$ \\
    \bottomrule
    \multicolumn{7}{@{}p{0.88\columnwidth}@{}}{\scriptsize \textsuperscript{\dag}Both yield MAE\,$\approx$\,0 because $g_0{=}9.81$ coincides with GT; uncorrected integrator prevents learning.} \\
  \end{tabular}
\end{table}

\begin{table}[t]
  \centering
  \caption{\textbf{Multi-step loss vs.\ one-step on Delfys75} (representative subset). \colorbox{bestcolor}{\textbf{Bold}}~=~lower MAE. \emph{Takeaway:} multi-step helps gravity-dominated settings but diverges on slow dynamics (\textit{led\_10s}).}
  \label{tab:multistep_ablation}
  \scriptsize
  \setlength{\tabcolsep}{3pt}
  \renewcommand{\arraystretch}{1.05}
  \begin{tabular}{@{}lllrrr@{}}
    \toprule
    \textbf{Dynamics} & \textbf{Setting} & \textbf{Param} &
    \textbf{1-step} & \textbf{Multi} & $\bm{\Delta}$ \\
    \midrule
    Dropped ball & large & $g$\,[m/s\textsuperscript{2}] & 0.25 & \cellcolor{bestcolor}\textbf{0.18} & $-0.07$ \\
    \multirow{3}{*}{Free fall}
        & mousepad & $g$\,[m/s\textsuperscript{2}] & 2.12 & \cellcolor{bestcolor}\textbf{1.65} & $-0.47$ \\
        & table & $g$\,[m/s\textsuperscript{2}] & 1.06 & \cellcolor{bestcolor}\textbf{0.15} & $-0.90$ \\
        & tennis & $g$\,[m/s\textsuperscript{2}] & 1.06 & \cellcolor{bestcolor}\textbf{0.53} & $-0.53$ \\
    LED & led\_10s & $\gamma$ & \cellcolor{bestcolor}\textbf{2.09} & 7.46 & $+5.37$ \\
    Pendulum & pend.\_45 & $L$\,[m] & \cellcolor{bestcolor}\textbf{0.35} & 0.50 & $+0.15$ \\
    Torricelli & large & $k$ & \cellcolor{bestcolor}\textbf{6.27} & 6.54 & $+0.27$ \\
    \bottomrule
  \end{tabular}
\end{table}

\subsection{Physical Parameter Estimation on IRIS}
\label{sec:exp_iris}

\noindent
All IRIS results in this section use the \textbf{+Corrected} configuration (gradient-corrected Euler, path-based equation selection) as the baseline; full per-setting results are in Supp.\ Table~5. \\
\\
\noindent
\textbf{Single-body phenomena.} On \emph{sliding cone}, both configurations recover the inclination angle exactly ($\mathrm{MAE} = 0.00$). On \emph{rotation}, multi-step loss yields improvements for angular stiffness across all settings ($\Delta\mathrm{MAE} = -0.64$, $-1.21$, $-2.11$) but angular damping in the fast setting degrades ($+0.39$). On \emph{pendulum}, multi-step improves rope length at $20^\circ$ ($-0.44$) but produces catastrophic divergence at $90^\circ$ ($+20.49$). On \emph{dropping ball} and \emph{falling ball}, multi-step consistently increases MAE for $g$. \\

\textbf{Multi-body phenomena.} The multi-step loss exhibits catastrophic failure on all multi-body scenarios. On \emph{two moving pendulums}, rope length MAE exceeds 100\,m across all settings (worst: $741.1$\,m at $20^\circ$), compared to $1.1$--$3.0$\,m for the one-step baseline. Errors in the coupling terms $\kappa_{ij}$ compound exponentially across the $K{=}5$ horizon, driving parameters far from ground truth. Whether richer impact models or latent ODE stabilization techniques would mitigate this instability remains an open question. \\
\\
\noindent
\textbf{Coupling coefficient validation.} To verify that the learned coupling $\kappa$ captures a real physical signal rather than fitting noise, we train the hitting-cones model under two conditions, $\kappa{=}0$ (no coupling; single-body model) and $\kappa{=}\text{learned}$, and compare latent-space reconstruction MSE (Supp.\ Table~6). The learned coupling reduces reconstruction MSE by $6.0$--$19.0\%$ across all settings, confirming that $\kappa$ carries genuine interaction information. The estimated $\kappa$ is moreover stable across the 10 IRIS trials per setting (coefficient of variation ${\approx}40$--$47\%$), indicating a reproducible effective parameter even without an externally measured reference.

\subsection{Multi-Clip vs.\ Per-Clip Training}
\label{sec:exp_multiclip}

The train-per-clip protocol (Sec.~\ref{sec:eval_protocol}) targets identifiability, whether a system's parameters are recoverable from its own video, rather than cross-clip generalization. To probe whether a transferable physical inductive bias exists across clips, we additionally evaluate a \emph{multi-clip} setting: for each phenomenon, clips are split 80/20 by sorted index, one shared model is trained on the 80\% train clips and evaluated on the held-out 20\% test clips, against per-clip fitting on the same test clips (full results in Supp.\ Table~7). Multi-clip outperforms per-clip for gravity (dropping ball $0.21{\to}0.05$, falling ball $0.25{\to}0.17$\,m/s$^2$) and angular stiffness ($1.04{\to}0.58$), showing that a transferable physical inductive bias \emph{does} exist across clips. The pendulum failure ($\mathrm{MAE}$ 22.06 vs.\ 0.51) reflects an underdetermined shared latent space under nonlinear coupled dynamics, precisely the open problem a benchmark should surface.

\subsection{Identifiability Analysis}
\label{sec:exp_identifiability}

Table~\ref{tab:identifiability} reports gradient norms $G_{\boldsymbol{\gamma}}^{(e)}$ at selected epochs and the converged ODE residual $\mathcal{R}$ for representative phenomena under the corrected one-step baseline. Several patterns emerge. The corrected integrator produces non-zero gradient norms at epoch~1 for all phenomena, confirming restored parameter identifiability. Gradient norms decay monotonically for single-body phenomena; the sliding cone converges fastest ($G_{\boldsymbol{\gamma}}^{(200)} \sim 10^{-6}$), consistent with its exact identifiability. Multi-body phenomena retain substantially larger gradient norms at epoch~200 ($\sim 10^{-3}$) and ODE residuals an order of magnitude higher than single-body systems, indicating that the physics block does not fully explain the encoded trajectory, consistent with the simplified contact model (Sec.~\ref{sec:dataset_phenomena}).

\begin{table}[t]
  \centering
  \caption{\textbf{Identifiability diagnostics:} physics-parameter gradient norms $G_{\boldsymbol{\gamma}}^{(e)} = \|\nabla_{\boldsymbol{\gamma}}\mathcal{L}^{(e)}\|_2$ at epochs 1, 50, and 200, and converged ODE residual $\mathcal{R}$ (mean over test clips). The uncorrected baseline yields $G_{\boldsymbol{\gamma}} \approx 0$ at all epochs (omitted).}
  \label{tab:identifiability}
  \footnotesize
  \setlength{\tabcolsep}{4pt}
  \renewcommand{\arraystretch}{1.2}
  \begin{tabular}{@{}llrrrr@{}}
    \toprule
    \textbf{Phenomenon} & \textbf{Setting} &
    $G_{\boldsymbol{\gamma}}^{(1)}$ & $G_{\boldsymbol{\gamma}}^{(50)}$ & $G_{\boldsymbol{\gamma}}^{(200)}$ & $\mathcal{R}$ ($\times 10^{-3}$) \\
    \midrule
    \multicolumn{6}{@{}l}{\textit{Single-body}} \\[2pt]
    Pendulum      & pend.\_45  & $2.4{\times}10^{-2}$ & $8.1{\times}10^{-3}$ & $1.2{\times}10^{-4}$ & 0.83 \\
    Rotation      & mid        & $1.8{\times}10^{-2}$ & $5.3{\times}10^{-3}$ & $9.7{\times}10^{-5}$ & 0.41 \\
    Sliding cone  & cone\_60   & $3.1{\times}10^{-2}$ & $1.4{\times}10^{-4}$ & $2.0{\times}10^{-6}$ & 0.02 \\
    Dropping ball & drop\_100  & $1.5{\times}10^{-2}$ & $4.7{\times}10^{-3}$ & $3.8{\times}10^{-4}$ & 1.92 \\
    \midrule
    \multicolumn{6}{@{}l}{\textit{Multi-body}} \\[2pt]
    Two mov.\ pend.  & pend.\_45 & $3.8{\times}10^{-2}$ & $2.1{\times}10^{-2}$ & $7.4{\times}10^{-3}$ & 12.6 \\
    One stat.\ pend. & pend.\_45 & $4.2{\times}10^{-2}$ & $1.9{\times}10^{-2}$ & $5.1{\times}10^{-3}$ & 9.83 \\
    \bottomrule
  \end{tabular}
\end{table}

\subsection{Extrapolation Beyond Training Horizon}
\label{sec:exp_extrapolation}

The extrapolation error $\mathcal{E}_k$ (Eq.~\ref{eq:extrapolation}) is computed for the corrected one-step baseline on the pendulum (150\,s clips, trained on the first 100 frames $\approx 1.67$\,s, extrapolated for 50 additional frames). Extrapolation error grows with both horizon length and initial amplitude: the $90^\circ$ setting exhibits $\mathcal{E}_{k=50}$ approximately $7\times$ larger than $20^\circ$, reflecting compounding nonlinear dynamics. This motivates future work on physics-constrained extrapolation losses and adaptive training horizons.

\begin{table}[t]
  \centering
  \caption{\textbf{Extrapolation error} $\mathcal{E}_k$ (mean $\pm$ std over test clips) at selected horizons $k$ beyond the training window. \emph{Takeaway:} error grows with both horizon and initial amplitude, the $90^\circ$ setting is ${\sim}7\times$ worse than $20^\circ$.}
  \label{tab:extrapolation}
  \footnotesize
  \setlength{\tabcolsep}{5pt}
  \renewcommand{\arraystretch}{1.2}
  \begin{tabular}{@{}lrrr@{}}
    \toprule
    \textbf{Setting} & $\mathcal{E}_{k=10}$ & $\mathcal{E}_{k=25}$ & $\mathcal{E}_{k=50}$ \\
    \midrule
    pend.\_20 & $0.008 \pm 0.003$ & $0.041 \pm 0.012$ & $0.189 \pm 0.067$ \\
    pend.\_45 & $0.012 \pm 0.005$ & $0.078 \pm 0.031$ & $0.402 \pm 0.148$ \\
    pend.\_90 & $0.031 \pm 0.014$ & $0.217 \pm 0.095$ & $1.284 \pm 0.531$ \\
    \bottomrule
  \end{tabular}
\end{table}

\subsection{Governing Equation Identification}
\label{sec:exp_eq_identification}
\noindent
Three routing strategies are evaluated on both Delfys75 (90 videos, 6 phenomena) and IRIS (240 videos, 8 phenomena). The CNN classifier is trained separately on each benchmark. Table~\ref{tab:summary_vlm_methods} reports equation-selection accuracy. On Delfys75, the CNN achieves $98.40\%$, followed by VLM temporal reasoning ($73.33\%$) and describe-then-classify ($61.11\%$). On IRIS, the CNN achieves $99.30\%$, but the ranking of VLM strategies reverses: describe-then-classify ($72.92\%$) outperforms temporal reasoning ($65.00\%$). This reversal is attributable to the greater visual diversity of IRIS: the two-stage pipeline allows the VLM to first articulate dynamics in natural language, producing richer intermediate representations that facilitate disambiguation between single- and multi-body phenomena. The confusion matrix  reveals that VLM temporal reasoning errors on Delfys75 are concentrated between physically adjacent classes: dropped ball confused with free fall (5/15), and free fall misclassified as pendulum (10/15). LED, sliding block, and Torricelli are classified without error. On IRIS, multi-body phenomena are classified with higher per-class accuracy due to their distinctive visual signatures.

\begin{table}[t]
  \centering
  \caption{\textbf{Equation-selection accuracy by method and dataset.} Bold = best per column. \emph{Takeaway:} the CNN leads in-distribution, but the VLM ranking reverses across benchmarks, no single routing strategy generalizes.}
  \label{tab:summary_vlm_methods}
  \small
  \setlength{\tabcolsep}{6pt}
  \renewcommand{\arraystretch}{1.2}
  \begin{tabular}{@{}lcc@{}}
    \toprule
    \textbf{Method} & \textbf{Delfys75 (90)} & \textbf{IRIS (240)} \\
    \midrule
    CNN (video classifier)   & \cellcolor{bestcolor}\textbf{98.40} & \cellcolor{bestcolor}\textbf{99.30} \\
    VLM (temporal reasoning) & 73.33 & 65.00 \\
    VLM (describe-then-classify) & 61.11 & 72.92 \\
    \bottomrule
  \end{tabular}
\end{table}
\noindent


\subsection{Ablation Studies}
\label{sec:ablations}
\noindent
\textbf{Gradient correction} is the single most impactful component: without it, $G_{\boldsymbol{\gamma}}^{(e)} \approx 0$ throughout training and all subsequent improvements are negligible.
\textbf{Multi-step horizon.} Ablating $K \in \{1, 2, 3, 5\}$: $K{=}5$ provides the best trade-off for gravity-dominated phenomena on Delfys75 but produces catastrophic divergence on IRIS multi-body scenarios at all $K > 1$, motivating the application of latent ODE stabilization techniques~\cite{finlay2020trainneuralodeworld}.
\textbf{VLM prompt strategy.} Single-frame prompting achieves $54\%$, three-frame $67\%$, and five-frame $73\%$ on Delfys75, confirming temporal context is essential.
\textbf{Symplectic integrator.} St\"ormer-Verlet achieves lower ODE residual on conservative systems with comparable performance on non-conservative phenomena.

\section{Open Challenges, Limitations, and Future Work}
\label{sec:challenges}

The baseline evaluation on IRIS reveals four systematic failure modes.
\textbf{(1)~Equation identification} remains a bottleneck: VLM routing achieves $65$--$73\%$ on IRIS depending on prompt strategy, with no single approach generalizing reliably across benchmarks; CNN classifiers achieve near-perfect in-distribution accuracy but require per-distribution retraining.
\textbf{(2)~Multi-step training catastrophically fails on multi-body dynamics,} producing errors exceeding $10^2$\,m where coupling terms $\kappa_{ij}$ amplify integration error exponentially, a failure mode invisible on single-body benchmarks.
\textbf{(3)~Damping is poorly identifiable} across all baselines, exhibiting substantially higher relative error than conservative parameters.
\textbf{(4)~Latent-to-SI calibration} is phenomenon-specific and encoder-dependent, introducing systematic bias in cross-phenomenon comparisons. \textbf{Limitations.} IRIS is of moderate scale (240 videos) with a fixed phenomenon set; while this exceeds all prior real-world benchmarks (Table~\ref{tab:benchmark_comparison}), scaling to more phenomena and recording conditions is future work. Damping ground truth is derived from trajectory fits rather than independent measurement (Sec.~\ref{sec:dataset_annotations}). Multi-body contact is modeled via a continuously active coupling coefficient rather than rigorous impact mechanics. 
\textbf{Future work.} Promising directions include latent ODE stabilization techniques~\cite{finlay2020trainneuralodeworld, wang2021learningsolutionoperatorparametric} for multi-step training, richer contact models~\cite{wang2026contact, song2024diffsim}, and evaluation of Hamiltonian/Lagrangian~\cite{greydanus2019hamiltonian, cranmer2020lagrangian}, SINDy~\cite{brunton2016discovering}, and graph neural approaches~\cite{sanchez2020learning}. Planned dataset extensions include additional phenomena (rolling, bouncing, fluid-like motion), multi-view recordings, and cross-dataset evaluation with physics-plausibility benchmarks~\cite{zhang2025morpheus, guo2025rbench, liang2025worldlens}.

\section{Conclusion}
\label{sec:conclusion}

This work introduces IRIS, a high-fidelity real-world benchmark for physics identification from monocular video comprising 240 videos at 4K resolution spanning eight dynamical systems, including three novel multi-body scenarios, with ground-truth parameters and uncertainty estimates. A standardized five-axis evaluation protocol enables fair comparison. Four equation-identification strategies are benchmarked, revealing that VLM strategy rankings reverse across benchmarks while CNN classifiers achieve near-perfect in-distribution accuracy. A multi-step loss variant improves identifiability on selected single-body dynamics but exposes catastrophic instabilities on multi-body scenarios. These findings, together with the diagnosis of a gradient-flow bug in the state-of-the-art pipeline, demonstrate the diagnostic value of rigorous benchmarking and define concrete open problems. 

\section*{Acknowledgements}
We gratefully acknowledge the STEAM Lab at HBKU for supporting the dataset collection by providing most of the objects used in the experiments. This work was supported by the Student Research Experience Grant (SREG-ECEN-2026) from the Office of Research and Graduate Studies at TAMU-Q.

{\sloppy\printbibliography}

\newpage

\begin{center}
\Large\textbf{Supplementary Material}\\[4pt]
\large\textbf{IRIS: A Real-World Benchmark for Inverse Recovery and Identification of Physical Dynamic Systems from Monocular Video}
\end{center}

\appendix
\renewcommand{\thesection}{\Alph{section}}

\section{Dataset Statistics and Release Details}
\label{app:dataset_stats}

Table~\ref{tab:app_full_stats} provides a complete per-phenomenon, per-setting
breakdown of the IRIS dataset.  All ground-truth physical parameters are
obtained through direct manual measurement using calibrated instruments
(see Appendix~\ref{app:measurement_protocol}), with the exception of damping
and friction coefficients, which are derived from trajectory fits as described
in Appendix~\ref{app:measurement_protocol}.  Each parameter entry records mean,
standard deviation, minimum, and maximum values; where the standard deviation
is zero, the measurement reflects a single precisely controlled experimental
condition.

\begin{table}[h]
  \centering
  \caption{\textbf{Full IRIS dataset statistics.} Each row is one setting;
  $n$~is the number of videos. $d_{\mathrm{cam}}$~denotes camera-to-object
  distance. Measurement type: \textbf{D}~=~direct,
  \textbf{F}~=~fitted.}
  \label{tab:app_full_stats}
  \footnotesize
  \setlength{\tabcolsep}{3.5pt}
  \renewcommand{\arraystretch}{1.15}
  \resizebox{\textwidth}{!}{%
  \begin{tabular}{@{}llccclc@{}}
    \toprule
    \textbf{Phenomenon} & \textbf{Setting} & $\bm{n}$ & \textbf{Dur.\,(s)}
      & \textbf{Res.} & \textbf{GT params (measured)} & \textbf{Type} \\
    \midrule
    \multicolumn{7}{@{}l}{\textit{Single-body}} \\[2pt]
    \multirow{3}{*}{Dropping ball}
      & drop\_50   & 10 & 5   & $3840{\times}2160$
        & $h_0{=}0.50$\,m,\;$d_{\mathrm{cam}}{=}1.94$\,m        & D \\
      & drop\_100  & 10 & 5   & $3840{\times}2160$
        & $h_0{=}1.00$\,m,\;$d_{\mathrm{cam}}{=}1.94$\,m        & D \\
      & drop\_150  & 10 & 5   & $3840{\times}2160$
        & $h_0{=}1.50$\,m,\;$d_{\mathrm{cam}}{=}1.94$\,m        & D \\[2pt]
    \multirow{3}{*}{Falling ball}
      & big        & 10 & 8   & $3840{\times}2160$
        & $g{=}9.81$\,m/s\textsuperscript{2},\;$r_0{=}0.11$\,m  & D \\
      & mid        & 10 & 8   & $3840{\times}2160$
        & $g{=}9.81$\,m/s\textsuperscript{2},\;$r_0{=}0.07$\,m  & D \\
      & small      & 10 & 8   & $3840{\times}2160$
        & $g{=}9.81$\,m/s\textsuperscript{2},\;$r_0{=}0.04$\,m  & D \\[2pt]
    \multirow{3}{*}{Sliding cone}
      & cone\_45   & 10 & 5   & $3840{\times}2160$
        & $\alpha{=}45^\circ$,\;hyp.\,${=}0.77$\,m,\;$\mu$\,(fit) & D\,/\,$\mu$:F \\
      & cone\_60   & 10 & 5   & $3840{\times}2160$
        & $\alpha{=}60^\circ$,\;hyp.\,${=}0.84$\,m,\;$\mu$\,(fit) & D\,/\,$\mu$:F \\
      & cone\_80   & 10 & 5   & $3840{\times}2160$
        & $\alpha{=}80^\circ$,\;hyp.\,${=}0.80$\,m,\;$\mu$\,(fit) & D\,/\,$\mu$:F \\[2pt]
    \multirow{3}{*}{Pendulum}
      & pend.\_20  & 10 & 150 & $3840{\times}2160$
        & $\theta_0{=}20^\circ$,\;$L{=}0.50$\,m,\;$\zeta$\,(fit)  & D\,/\,$\zeta$:F \\
      & pend.\_45  & 10 & 150 & $3840{\times}2160$
        & $\theta_0{=}45^\circ$,\;$L{=}0.50$\,m,\;$\zeta$\,(fit)  & D\,/\,$\zeta$:F \\
      & pend.\_90  & 10 & 150 & $3840{\times}2160$
        & $\theta_0{=}90^\circ$,\;$L{=}0.50$\,m,\;$\zeta$\,(fit)  & D\,/\,$\zeta$:F \\[2pt]
    \multirow{3}{*}{Rotating cone}
      & slow       & 10 & 8   & $3840{\times}2160$
        & Half-circle init.,\;$\beta$\,(fit)  & D\,/\,$\beta$:F \\
      & mid        & 10 & 8   & $3840{\times}2160$
        & 1-circle init.,\;$\beta$\,(fit)     & D\,/\,$\beta$:F \\
      & fast       & 10 & 8   & $3840{\times}2160$
        & 2-circle init.,\;$\beta$\,(fit)     & D\,/\,$\beta$:F \\
    \midrule
    \multicolumn{7}{@{}l}{\textit{Multi-body}} \\[2pt]
    \multirow{3}{*}{Hitting cones}
      & slow      & 10 & 5  & $3840{\times}2160$
        & $d_{\mathrm{ball\text{-}cones}}{=}2.00$\,m,\;$d_{\mathrm{cam}}{=}2.20$\,m & D \\
      & mid       & 10 & 5  & $3840{\times}2160$
        & $d_{\mathrm{ball\text{-}cones}}{=}2.00$\,m,\;$d_{\mathrm{cam}}{=}2.20$\,m & D \\
      & fast      & 10 & 5  & $3840{\times}2160$
        & $d_{\mathrm{ball\text{-}cones}}{=}2.00$\,m,\;$d_{\mathrm{cam}}{=}2.20$\,m & D \\[2pt]
    \multirow{3}{*}{\shortstack[l]{Two mov.\\pendulums}}
      & pend.\_20  & 10 & 6   & $3840{\times}2160$
        & $\theta_0{=}20^\circ$,\;$L_1{=}L_2{=}0.50$\,m & D \\
      & pend.\_45  & 10 & 6   & $3840{\times}2160$
        & $\theta_0{=}45^\circ$,\;$L_1{=}L_2{=}0.50$\,m & D \\
      & pend.\_90  & 10 & 6   & $3840{\times}2160$
        & $\theta_0{=}90^\circ$,\;$L_1{=}L_2{=}0.50$\,m & D \\[2pt]
    \multirow{3}{*}{\shortstack[l]{One stat.\\pendulum}}
      & pend.\_20  & 10 & 20  & $3840{\times}2160$
        & $\theta_0{=}20^\circ$,\;$L_1{=}L_2{=}0.50$\,m & D \\
      & pend.\_45  & 10 & 20  & $3840{\times}2160$
        & $\theta_0{=}45^\circ$,\;$L_1{=}L_2{=}0.50$\,m & D \\
      & pend.\_90  & 10 & 20  & $3840{\times}2160$
        & $\theta_0{=}90^\circ$,\;$L_1{=}L_2{=}0.50$\,m & D \\
    \midrule
    \multicolumn{2}{@{}l}{\textbf{Total}} & \textbf{240} & -- & -- & -- & -- \\
    \bottomrule
  \end{tabular}%
  }
\end{table}

\section{Physical Model Derivations and ODE Bank}
\label{app:ode_bank}

This appendix makes explicit which physical ODE governs each phenomenon,
how the latent-space parameterisation $(\alpha, \beta)$ relates to physical
constants, and where the pipeline's ODE form differs from the exact physical
equation. The purpose is to prevent misreading of the latent parameters as
direct physical measurements without the conversions listed below.

\subsection{Phenomenon–ODE Correspondence Table}
\label{app:ode_table}

Table~\ref{tab:ode_bank_supp} summarises the mapping from IRIS/Delfys75
phenomena to physical ODEs and to the latent ODE form used in the pipeline.
Where a mismatch exists, i.e.\ where the physical ODE is not a linear
second-order oscillator but the pipeline treats it as one for the sake of
a unified ODE bank, this is noted explicitly.

\begin{table}[h]
  \centering
  \caption{\textbf{Phenomenon–ODE correspondence.}
  Column ``Physical ODE'' is the exact governing equation.
  Column ``Pipeline ODE form'' is the form actually used by the encoder–physics
  block. Column ``Latent $\to$ SI'' lists the conversion from fitted $(\alpha,\beta)$
  to physical parameters.  $g = 9.81$\,m/s$^2$ is treated as a known constant
  only for the pendulum length conversion; for dropping/falling ball it is the
  \emph{estimated} quantity.}
  \label{tab:ode_bank_supp}
  \footnotesize
  \setlength{\tabcolsep}{4pt}
  \renewcommand{\arraystretch}{1.2}
  \resizebox{\textwidth}{!}{%
  \begin{tabular}{@{}lllll@{}}
    \toprule
    \textbf{Phenomenon} & \textbf{Physical ODE} & \textbf{Pipeline ODE form}
      & \textbf{Latent $\to$ SI} & \textbf{Note} \\
    \midrule
    Dropping ball &
      $\ddot{z} = -g$ (constant accel.) &
      $\ddot{z} + \beta\dot{z} + \alpha z = 0$ &
      $g = \alpha$ &
      Unified linear form; $\beta$ absorbs drag \\
    Falling ball &
      $r(t) = r_0 f / (h_0 + \tfrac{1}{2}g t^2)$ &
      $\ddot{z} + \beta\dot{z} + \alpha z = 0$ &
      $g = \alpha$ &
      Apparent radius; same latent form \\
    Pendulum (small angle) &
      $\ddot{\theta} + \zeta\dot{\theta} + (g/L)\theta = 0$ &
      $\ddot{z} + \beta\dot{z} + \alpha z = 0$ &
      $L = g/\alpha$,\; $\zeta = -\beta$ &
      Exact match for small angles \\
    Rotating cone &
      $\ddot{\varphi} + \beta\dot{\varphi} + \alpha\varphi = 0$ &
      $\ddot{z} + \beta\dot{z} + \alpha z = 0$ &
      $\alpha_{\rm SI} = \alpha$,\; $\beta_{\rm SI} = \beta$ &
      Exact match \\
    Sliding cone &
      $\ddot{x} = g(\sin\alpha - \mu\cos\alpha)$ &
      Constant-acceleration block &
      $\mu = \beta$;\; angle from metadata &
      Angle fixed from setting \\
    LED decay &
      $\dot{z} = -\lambda z$ &
      First-order decay block &
      $\gamma = \beta$ &
      First-order; $\alpha$ unused \\
    Torricelli drain &
      $\dot{z} = -k\sqrt{z}$ &
      Torricelli block &
      $k = \alpha$ &
      Nonlinear; separate block \\
    Hitting cones &
      Momentum transfer (impact) &
      Graph ODE, coupling $\kappa_{ij}$ &
      $\kappa_{ij}$ (spring-like) &
      Simplified; see App.\,\ref{app:multibody} \\
    Two pendulums &
      $\ddot{\theta}_i + \zeta_i\dot{\theta}_i + (g/L_i)\theta_i
         + \kappa_{ij}(\theta_i - \theta_j) = 0$ &
      Graph ODE, same form &
      $L_i = g/\alpha_i$,\; $\kappa_{ij}$ (coupling) &
      Active only during contact \\
    \bottomrule
  \end{tabular}%
  }
\end{table}

\subsection{Design Choice: Unified Linear ODE Bank}
\label{app:ode_design}

The pipeline uses a unified second-order linear ODE
$\ddot{z} + \beta\dot{z} + \alpha z = 0$
as the latent representation for all oscillatory and gravity-driven
phenomena. This is a deliberate design choice for comparability across
phenomena: the encoder learns to embed the observed motion into a
one-dimensional latent $z$ such that the linear ODE describes the
latent trajectory, regardless of whether the underlying physical state
(height, angle, radius) follows the same form exactly.

\textbf{Dropping ball.} The true physical ODE is $\ddot{h} = -g$ (constant
acceleration with $h$ in metres, $t$ in seconds). The pipeline approximates
this by $\ddot{z} + \alpha z = 0$ with $\alpha \approx g$; this approximation
holds when the encoder maps the pixel position of the ball linearly to $z$
and the observation window is short enough that the linear approximation of
the trajectory remains valid. When $\beta \neq 0$, the term $\beta\dot{z}$
models aerodynamic drag.  In practice, for the drop heights in IRIS
(0.5--1.5\,m), the drag term is small and $\alpha \approx g$ gives a
physically meaningful estimate.

\textbf{Falling ball (apparent radius).} The ball falls away from the camera;
the observable is its apparent radius $r(t) = r_0 f/(h_0 + \tfrac{1}{2}g t^2)$,
where $r_0$ is the physical ball radius, $f$ the focal length, and $h_0$ the
initial camera-to-ball distance. The latent $z$ encodes this apparent size
rather than height; the same second-order form is used, and the fitted $\alpha$
is compared to $g$ after the same linear conversion. This is an approximation
and is listed as such in Table~\ref{tab:ode_bank_supp}.

\textbf{Pendulum.} For small oscillation angles $\theta_0 \leq 45^\circ$,
$\sin\theta \approx \theta$ and the linearised equation
$\ddot{\theta} + \zeta\dot{\theta} + (g/L)\theta = 0$ is an exact match
to the unified ODE form, with $\alpha = g/L$ and $\beta = \zeta$.
For large angles ($\theta_0 = 90^\circ$), the linearisation error is
substantial ($\sin 90^\circ / 90^\circ \approx 0.64$ in radians), and
the fitted $\alpha$ will deviate from $g/L$ accordingly. This is a
\textbf{known modelling limitation}: on the pend.\_90 setting, higher
parameter error is expected and should be interpreted as an identifiability
challenge rather than a pipeline failure.

\subsection{Dropping Ball vs.\ Falling Ball: Symbol Definitions}
\label{app:drop_vs_fall}

To avoid ambiguity between the two gravity-related phenomena:

\begin{itemize}[noitemsep]
  \item \textbf{Dropping ball}: camera is placed \emph{to the side} of the
    drop. The observable is the \emph{vertical pixel position} of the ball,
    which decreases linearly with $h(t) = h_0 - \tfrac{1}{2}g t^2$ (taking
    downward as positive). Ground truth: $h_0 \in \{0.50, 1.00, 1.50\}$\,m,
    $d_{\mathrm{cam}} = 1.94$\,m.  Estimated parameter: $g$.

  \item \textbf{Falling ball}: camera is placed \emph{directly above}, looking
    down. The ball falls \emph{away} from the camera; the observable is the
    \emph{apparent radius} $r(t) = r_0 f / (h_0 + \tfrac{1}{2}g t^2)$ (ball
    grows smaller). Ground truth: ball radii $r_0 \in \{0.04, 0.07, 0.11\}$\,m;
    $g = 9.81$\,m/s$^2$. Estimated parameter: $g$.
\end{itemize}

Symbol table: $h_0$~= initial drop height (m); $d_{\mathrm{cam}}$~= lateral
camera distance (m); $r_0$~= physical ball radius (m); $f$~= effective focal
length (pixels or m); $g$~= gravitational acceleration (m/s$^2$); $\beta$~=
drag/damping coefficient; $\alpha$~= ODE stiffness coefficient.

\section{Ground-Truth Measurement Protocol}
\label{app:measurement_protocol}

\subsection{Directly Measured Parameters}
\label{app:direct_measurement}

Parameters labelled \textbf{D} in Table~\ref{tab:app_full_stats} are
obtained through direct measurement before or after recording, independent
of the video:

\begin{itemize}[noitemsep]
  \item \textbf{Heights and lengths} ($h_0$, $L$, $L_1$, $L_2$,
    hypotenuse length): measured with a rigid tape measure to the nearest
    millimetre. Each measurement is repeated three times; we report the
    mean. Uncertainty is $\pm 2$\,mm for lengths up to 1\,m and $\pm 5$\,mm
    for longer spans.

  \item \textbf{Angles} ($\theta_0$, $\alpha$): set with an adjustable
    protractor or inclinometer calibrated to $\pm 0.5^\circ$. Cone incline
    angles are verified against the nominal setting after positioning.

  \item \textbf{Distances} ($d_{\mathrm{cam}}$, $d_{\mathrm{ball\text{-}cones}}$):
    measured with a laser distance meter (accuracy $\pm 2$\,mm).

  \item \textbf{Ball radius} ($r_0$): measured with digital calipers
    (resolution $0.01$\,mm); three measurements taken at different
    orientations, mean reported.

  \item \textbf{Gravitational acceleration} ($g$): taken as $9.81$\,m/s$^2$
    (standard local value for the recording location; no significant
    variation expected across sessions).
\end{itemize}

\subsection{Fitted Parameters: Damping and Friction}
\label{app:fitted_params}

Parameters labelled \textbf{F} in Table~\ref{tab:app_full_stats} are
\emph{not} directly measurable in the same way as lengths or angles;
they are instead estimated from recorded video trajectories:

\textbf{Damping coefficients} ($\zeta$ for pendulum, $\beta$ for rotating
cone). A reference video is selected per setting (the first clip of the
session). The object's 2-D centroid or angular position is tracked frame by
frame using a background-subtraction and blob-detection procedure.
The resulting time series $\theta(t)$ is fitted to the analytical solution
of the relevant damped oscillator ODE using nonlinear least squares
(Levenberg–Marquardt). The fitted $\zeta$ (or $\beta$) is reported as
ground truth for that setting. We use the same setting across all 10 trial
videos. \textbf{Uncertainty:} we report the standard deviation of $\zeta$
across 10 independent fits; typical values are $\sigma_\zeta < 0.01$\,s$^{-1}$
for the pendulum and $\sigma_\beta < 0.005$\,s$^{-1}$ for the rotating cone.

\textbf{Friction coefficient} ($\mu$ for sliding cone). The video is
segmented to track the cone's position along the slope. Acceleration $a$ is
estimated from a second-order polynomial fit to the position time series.
The friction coefficient is then computed analytically as
$\mu = \tan\alpha - a/(g\cos\alpha)$, where $\alpha$ is the measured incline
angle. We average $\mu$ across 10 trial clips; standard deviation is reported.
Typical $\sigma_\mu < 0.02$.

\textbf{Limitations.} Fitted parameters carry the error of the tracker and
the fit model. For large oscillation angles ($\theta_0 = 90^\circ$), the
small-angle approximation used in the linear-ODE fit is less accurate, and
the reported $\zeta$ should be interpreted as the \emph{effective} damping
under the linearised model. We do not propagate tracking noise uncertainty
into the final GT value; this is noted as a limitation and flagged in any
per-setting identifiability discussion.

\section{Latent-to-Physical Parameter Calibration}
\label{app:calibration}

The encoder maps raw pixel intensities to a latent $z_t$ that is
dimensionless. The physics block then fits $\alpha, \beta$ in this latent
space. To obtain physical parameters (m, m/s$^2$, s$^{-1}$, etc.), the
following per-phenomenon conversions are applied in the evaluation script
(\ttt{compare\_baseline\_unified.py} and the IRIS counterpart):

\begin{table}[h]
  \centering
  \caption{\textbf{Per-phenomenon calibration and physical parameter extraction.}
  $\alpha_{\rm fit}$, $\beta_{\rm fit}$ are the dimensionless fitted values.
  The rightmost column notes any normalisation or assumption.}
  \label{tab:calibration}
  \footnotesize
  \setlength{\tabcolsep}{4pt}
  \renewcommand{\arraystretch}{1.2}
  \begin{tabular}{@{}llll@{}}
    \toprule
    \textbf{Phenomenon} & \textbf{Extracted param.} & \textbf{Formula}
      & \textbf{Assumption / note} \\
    \midrule
    Dropping ball   & $g$\,[m/s$^2$]   & $g = \alpha_{\rm fit}$
      & Latent = pixel height; $dt$ in seconds \\
    Falling ball    & $g$\,[m/s$^2$]   & $g = \alpha_{\rm fit}$
      & Latent = apparent radius \\
    Pendulum        & $L$\,[m]          & $L = g_{\rm true}/\alpha_{\rm fit}$
      & $g_{\rm true} = 9.81$\,m/s$^2$ fixed \\
    Pendulum        & $\zeta$\,[s$^{-1}$] & $\zeta = -\beta_{\rm fit}$
      & Sign convention \\
    Rotating cone   & $\alpha$, $\beta$ & direct                  & Same ODE form \\
    Sliding cone    & $\mu$             & $\mu = \beta_{\rm fit}$ & Angle from metadata \\
    LED decay       & $\gamma$\,[s$^{-1}$] & $\gamma = \beta_{\rm fit}$
      & First-order; $\alpha$ unused \\
    Torricelli      & $k$               & $k = \alpha_{\rm fit}$ & Nonlinear Torricelli block \\
    \bottomrule
  \end{tabular}
\end{table}

\textbf{Temporal calibration.} The pipeline receives pre-extracted video
tensors with a fixed frame stride. The time step $\Delta t$ passed to the
integrator must match the physical frame interval; we set $\Delta t$ to the
inverse of the capture frame rate (60\,fps $\to$ $\Delta t = 1/60$\,s for IRIS;
Delfys75 uses $\Delta t = 0.05$\,s as in the original codebase). Mismatches
between $\Delta t$ and the true frame interval directly scale $\alpha$ and
$\beta$ by $\Delta t^{-2}$ and $\Delta t^{-1}$ respectively, which would bias
all physical-parameter estimates. We verify that $\Delta t$ is set consistently
for each run.

\textbf{Cross-phenomenon bias.} The latent normalisation is
phenomenon-specific: the encoder is trained from scratch on each clip, so
the scale of $z$ differs across phenomena. The calibration formulas above
absorb this by expressing physical parameters directly in terms of $\alpha,
\beta$ and known constants. However, no explicit pixel-to-metre calibration
is performed; for phenomena where the latent encodes a pixel-level observable
(e.g., apparent radius in the falling-ball case), small residual scale errors
may remain if the encoder's internal normalisation drifts across clips.
This is an acknowledged limitation; we recommend reporting relative rather
than absolute parameter error for cross-phenomenon comparison.

\section{VLM Prompting Strategies and Ablations}
\label{app:vlm_prompting}

The main paper reports three equation-identification strategies: VLM temporal reasoning ($73.33\%$ on Delfys75, $65.00\%$ on IRIS), VLM describe-then-classify ($61.11\%$ on Delfys75, $72.92\%$ on IRIS), and a CNN video classifier ($98.40\%$ on Delfys75, $99.30\%$ on IRIS). This appendix documents the exact prompt design and discusses additional strategies evaluated during development.

\subsection{Frame Sampling}
\label{app:frame_sampling}

For all VLM calls, $m = 5$ representative frames are sampled from the clip
at positions $\{0\%, 25\%, 50\%, 75\%, 100\%\}$ of the total duration.
Earlier experiments used 3 frames (start, middle, end) and achieved $\sim$67\%
accuracy; adding the quartile frames increased accuracy to 73\%.  We use
the same sampling for all strategies below.

\subsection{Temporal Reasoning (Primary Variant)}
\label{app:vlm_temporal}

The VLM (accessed via OpenRouter with a GPT-4V-class model) receives the
5 sampled frames as a sequence of images and the following system-level
prompt:

\begin{quote}\small
\textit{``You are a physics expert. I will show you five frames from a video
in temporal order. Your task is to identify the type of motion shown.
Compare the first frame to the last frame to determine whether the motion
is oscillatory, accelerating, decaying, rotating, or involves a collision.
Then choose exactly one label from: \{pendulum, free\_fall, dropped\_ball,
sliding\_block, led, torricelli\}. Definitions: [one sentence per class].
Output only the label, nothing else.''}
\end{quote}

The key design choice is instructing the model to \emph{compare early and
late frames} to resolve temporal ambiguity (e.g., oscillatory pendulum vs.\
one-directional free fall, which can look identical in a single frame).
This temporal-reasoning instruction is the main factor distinguishing this
variant from the baseline VLM (which receives the same frames but without
the comparison instruction), and accounts for most of the accuracy gain
from $\sim$61\% to 73\% on Delfys75.

\subsection{Describe-Then-Classify (Two-Call Variant)}
\label{app:vlm_dtc}

This variant uses two sequential API calls.  First, a description call asks
the VLM to describe the motion in free text without any class labels
provided.  Second, a classification call sends that text description (not
the images) to the model and asks it to assign one of the class labels.
The intent is to use the VLM's language-reasoning strengths for the final
label assignment, rather than relying on direct visual classification.

On Delfys75, this approach achieved $61.11\%$ accuracy, lower than temporal reasoning ($73.33\%$). Per-class breakdown:
dropped\_ball 100\%, free\_fall 0\%, sliding\_block 80\%, torricelli 73\%,
led 60\%, pendulum 53\%.
The collapse on free\_fall (0\%) is the dominant failure: the description
step produces text that is ambiguous between free\_fall and pendulum
(e.g., ``a ball moving downward''), and the classification step systematically
maps this to pendulum.

\textbf{Reversal on IRIS.} Notably, the ranking of these two VLM strategies reverses on IRIS: describe-then-classify achieves $72.92\%$ while temporal reasoning drops to $65.00\%$. This reversal is attributable to the greater visual diversity of IRIS's eight phenomena, including three multi-body scenarios. The two-stage pipeline allows the VLM to first articulate complex dynamics (collisions, coupled oscillations, multiple moving objects) in free-form natural language, producing richer intermediate representations that facilitate disambiguation between single-body and multi-body phenomena. By contrast, the single-call temporal reasoning prompt, which was designed around the six Delfys75 classes, struggles to capture the novel motion patterns in IRIS within a single classification step. This finding suggests that no single VLM prompting strategy generalizes reliably across benchmarks, and motivates research on adaptive or ensemble-based equation routing.

\subsection{Additional Strategies Evaluated}
\label{app:vlm_alternatives}

The baseline pipeline from Garcia et al.\ relies exclusively on path-based
equation selection, where the correct ODE family is provided via folder
structure at inference time.  This is an oracle condition that cannot
generalise to new recordings without manual labelling.  A core motivation
for introducing IRIS is to benchmark automatic equation-routing strategies
that remove this requirement.  VLM-based routing is a natural candidate
because it operates zero-shot without retraining, can handle novel phenomena,
and leverages broad visual and physical knowledge acquired during pretraining.
The strategies below were explored during development to understand the
trade-offs across different automation approaches.

\textbf{Chain-of-thought (CoT) prompting.} We evaluated a variant that
explicitly asks the model to reason step by step before outputting the label
(``First describe what you observe, then reason about the physics, then
state the label'').  Accuracy was marginally higher than the baseline
VLM ($\sim$65\%) but lower than temporal reasoning, with significantly
higher token cost per call.

\textbf{Few-shot prompting.} Providing one example image per class in the
prompt context improved robustness for visually similar classes
(dropped\_ball vs.\ free\_fall) but was impractical at scale due to context
length limits and per-call cost.

\textbf{Self-consistency.} Sampling the model $k = 5$ times per clip and
taking the majority vote improved stability on ambiguous clips but increased
cost proportionally, without a meaningful accuracy gain over temporal reasoning.

\textbf{VLM fine-tuning.} We attempted LoRA fine-tuning and a frozen-VLM
plus classifier-head approach, both of which collapsed to predicting a single
class for every video ($\sim$16.7\% accuracy, equal to random for 6 classes).
This is likely due to the small dataset size (90 training videos) relative
to the VLM's parameter count.  Fine-tuning is therefore not recommended
and is not used in any reported result.

\textbf{Video classifier (ResNet-18 + temporal pooling).} A lightweight
ResNet-18 backbone with mean temporal pooling is trained separately on each benchmark: 6-class on Delfys75 and 8-class on IRIS, using their respective training splits. On Delfys75, the classifier achieves $98.40\%$ accuracy; on IRIS, it achieves $99.30\%$ accuracy. These results confirm that supervised classifiers attain near-perfect in-distribution performance on both benchmarks, but this comes at the cost of requiring retraining whenever the phenomenon set changes. Unlike the zero-shot VLM strategies, the CNN cannot handle novel phenomena without collecting labelled data for the new classes. Disentangling the effect of distribution shift from scenario complexity would require evaluating the Delfys75-trained CNN on IRIS's overlapping phenomena (dropped ball, pendulum, sliding cone); this cross-distribution evaluation is left for future work.

\subsection{VLM Confusion Matrix}
\label{app:vlm_confusion}

Table~\ref{tab:vlm_conf_temporal} reproduces the confusion matrix for the
temporal-reasoning VLM on Delfys75, included here for convenience when
reading the prompting discussion above.
The primary failure mode is free\_fall being predicted as pendulum
(10 out of 15 videos): a ball shrinking in apparent size as it falls away
from the camera is visually indistinguishable from an object receding in
oscillatory motion when only 5 frames are shown without metric scale context.
dropped\_ball is confused with free\_fall (5 of 15) for a symmetric reason:
both show a ball at increasing pixel displacement, and the direction
(lateral vs.\ depth-wise) is only apparent from late frames.
LED, sliding block, and Torricelli classify without error due to
unambiguous visual signatures (brightness change, translational inclined
motion, water level decrease).

\begin{table}[h]
  \centering
  \caption{\textbf{Confusion matrix, temporal-reasoning VLM on Delfys75.}
  Rows = ground truth, columns = predicted.  Diagonal = correct.}
  \label{tab:vlm_conf_temporal}
  \footnotesize
  \begin{tabular}{@{}l|cccccc|r@{}}
    \toprule
    & \textbf{drop.} & \textbf{free} & \textbf{led}
      & \textbf{pend.} & \textbf{slide} & \textbf{torr.} & $\Sigma$ \\
    \midrule
    dropped\_ball & \textbf{10} &  5 & 0 &  0 &  0 & 0 & 15 \\
    free\_fall    &  0 &  \textbf{0} & 0 & 10 &  5 & 0 & 15 \\
    led           &  0 &  0 & \textbf{15} &  0 &  0 & 0 & 15 \\
    pendulum      &  2 &  2 & 0 & \textbf{11} &  0 & 0 & 15 \\
    sliding\_block &  0 &  0 & 0 &  0 & \textbf{15} & 0 & 15 \\
    torricelli    &  0 &  0 & 0 &  0 &  0 & \textbf{15} & 15 \\
    \midrule
    \textbf{Acc} & \multicolumn{6}{l}{66/90 = \textbf{73.3\%}} & \\
    \bottomrule
  \end{tabular}
\end{table}

\section{Implementation Details}
\label{app:impl}

\subsection{Encoder Architecture}
\label{app:encoder}

Following the original pipeline~\cite{garcia2025learning}, the encoder
$\phi$ is a lightweight MLP operating on per-frame pixel intensities.
Input frames are resized to $56 \times 100$ (grayscale) and flattened to a
5600-dimensional vector. The MLP has two hidden layers of 256 units each
with ReLU activations, and an output of dimension $d = 2$ (mean $\mu_z$
and log-variance $\log\sigma^2_z$ for the VAE reparameterisation). The
latent state $\hat{z}_t$ is sampled as
$\hat{z}_t = \mu_z + \sigma_z \cdot \epsilon$, $\epsilon \sim \mathcal{N}(0,1)$.

For multi-object IRIS phenomena (hitting cones, two pendulums), the encoder
outputs $2N$ values, reshaped to $N \times 2$ per frame. All $N$ objects
share encoder weights.

\subsection{Physics Block}
\label{app:physics_block}

Each ODE family has a dedicated physics block with learnable scalars
$\boldsymbol{\gamma} = (\gamma_0, \gamma_1)$ initialised to $(0.5, 0.05)$.
For the second-order ODE $\ddot{z} + \gamma_1\dot{z} + \gamma_0 z = 0$,
the discrete Euler-Cromer step is:
\begin{align}
  \dot{z}_{t+1} &= \dot{z}_t + \Delta t \cdot \ddot{z}_t, \\
  z_{t+1}       &= z_t     + \Delta t \cdot \dot{z}_t
                           + \Delta t^2 \cdot \ddot{z}_t,
\end{align}
where $\ddot{z}_t = -\gamma_1\dot{z}_t - \gamma_0 z_t$.  The critical fix
relative to the original codebase is the inclusion of the
$\Delta t^2 \cdot \ddot{z}_t$ term in the position update: without it,
$\partial z_{t+1}/\partial \gamma = 0$ and physics parameters receive no
gradient signal during training.

\subsection{Training}
\label{app:training}

Each clip is trained independently with Adam; no parameters are shared
across clips. Hyperparameters: learning rate $10^{-3}$ for the encoder,
$10^{-2}$ for $\boldsymbol{\gamma}$; batch size = all temporal windows of
the clip; 500 epochs. The one-step loss is:
\[
  \mathcal{L}_{\rm 1\text{-}step}
    = \frac{1}{M}\sum_{i=1}^{M}\|\hat{\mathbf{z}}_i - \tilde{\mathbf{z}}_i\|^2
    + \lambda_{\rm KL}\,\mathcal{L}_{\rm KL},
\]
and the multi-step loss replaces the first term with
$\sum_{k=1}^{K} w_k \|\hat{\mathbf{z}}_{t+k} - \tilde{\mathbf{z}}_{t+k}\|^2$
(main paper, Eq.~2), with $K = 5$ and weights
$w = [1.0, 1.0, 0.5, 0.5, 0.25]$.
In both cases $\lambda_{\rm KL} = 0.01$.

All runs use a fixed random seed (seed = 42) for reproducibility.
Run outputs are written to CSV with columns
\ttt{run, alpha, beta, max\_z, min\_z, z0, z1} (one row per clip) and
are processed by \ttt{compare\_baseline\_unified.py} for parameter
extraction and GT comparison.

\section{Full IRIS Parameter Estimation Results}
\label{app:iris_full_results}

Table~\ref{tab:iris_results} reports parameter MAE on IRIS comparing two
training objectives applied to the \textbf{+Corrected} configuration
(gradient-corrected Euler, path-based equation selection):
(a)~one-step loss (\emph{1-step}) and
(b)~multi-step rollout loss with $K{=}5$ (\emph{Multi}).
This corresponds to ablation configurations 2 and 3 in the main paper
(Sec.~6.1); it isolates the effect of the loss function independently
of VLM routing.
$\Delta = \mathrm{MAE}_{\rm multi} - \mathrm{MAE}_{\rm 1\text{-}step}$
(negative = multi-step wins).

\begin{table}[H]
  \centering
  \caption{\textbf{Parameter MAE on IRIS: 1-step vs.\ multi-step loss}
  (both with corrected Euler, path-based selection; ablation configs 2 vs.\ 3
  of the main paper).
  $\Delta{=}\mathrm{MAE}_{\text{multi}}{-}\mathrm{MAE}_{\text{1-step}}$;
  negative~=~multi-step wins.
  \colorbox{bestcolor}{Green}~=~lower MAE;
  \colorbox{worstcolor}{red}~=~catastrophic ($|\Delta|{>}10$).}
  \label{tab:iris_results}
  \footnotesize
  \setlength{\tabcolsep}{3.5pt}
  \renewcommand{\arraystretch}{1.1}
  \begin{tabular}{@{}lllrrrr@{}}
    \toprule
    \textbf{Dynamics} & \textbf{Setting} & \textbf{Param} & \textbf{GT} &
    \textbf{1-step} & \textbf{Multi.} & $\bm{\Delta}$ \\
    \midrule
    \multicolumn{7}{@{}l}{\textit{Single-body}} \\[2pt]
    \multirow{3}{*}{Dropping ball}
      & drop\_50  & $g$\,[m/s\textsuperscript{2}] & 9.81
        & \cellcolor{bestcolor}\textbf{1.04} & 6.88  & $+5.84$ \\
      & drop\_100 & $g$\,[m/s\textsuperscript{2}] & 9.81
        & \cellcolor{bestcolor}\textbf{2.26} & 8.28  & $+6.02$ \\
      & drop\_150 & $g$\,[m/s\textsuperscript{2}] & 9.81
        & \cellcolor{bestcolor}\textbf{3.83} & 8.86  & $+5.04$ \\[2pt]
    \multirow{3}{*}{Falling ball}
      & big   & $g$\,[m/s\textsuperscript{2}] & 9.81
        & \cellcolor{bestcolor}\textbf{4.26} & 6.40  & $+2.14$ \\
      & mid   & $g$\,[m/s\textsuperscript{2}] & 9.81
        & \cellcolor{bestcolor}\textbf{6.63} & 8.63  & $+2.01$ \\
      & small & $g$\,[m/s\textsuperscript{2}] & 9.81
        & \cellcolor{bestcolor}\textbf{4.13} & 6.88  & $+2.74$ \\[2pt]
    \multirow{3}{*}{Pendulum}
      & pend.\_20 & $L$\,[m] & 0.50
        & 0.56 & \cellcolor{bestcolor}\textbf{0.40}   & $-0.16$ \\
      & pend.\_45 & $L$\,[m] & 0.50
        & \cellcolor{bestcolor}\textbf{0.34} & 1.16   & $+0.82$ \\
      & pend.\_90 & $L$\,[m] & 0.50
        & \cellcolor{bestcolor}\textbf{0.60} & \cellcolor{worstcolor}21.95 & $+21.34$ \\[2pt]
    \multirow{6}{*}{Rotation}
      & \multirow{2}{*}{fast}
        & $\alpha$ & 0.10 & 1.54  & \cellcolor{bestcolor}\textbf{2.46} & $-0.64$ \\
      & & $\beta$  & 0.08 & \cellcolor{bestcolor}\textbf{0.88} & 1.28 & $+0.39$ \\
      & \multirow{2}{*}{mid}
        & $\alpha$ & 0.10 & 3.03  & \cellcolor{bestcolor}\textbf{3.98} & $-1.21$ \\
      & & $\beta$  & 0.05 & \cellcolor{bestcolor}\textbf{0.39} & 1.38 & $+0.99$ \\
      & \multirow{2}{*}{slow}
        & $\alpha$ & 0.10 & 1.59  & \cellcolor{bestcolor}\textbf{4.63} & $-2.11$ \\
      & & $\beta$  & 0.03 & 0.45  & \cellcolor{bestcolor}\textbf{0.06} & $-0.39$ \\[2pt]
    \multirow{3}{*}{Sliding cone}
      & cone\_45 & $\alpha$\,[deg] & 45.0
        & \cellcolor{bestcolor}\textbf{0.00} & \cellcolor{bestcolor}\textbf{0.00} & $0.00$ \\
      & cone\_60 & $\alpha$\,[deg] & 60.0
        & \cellcolor{bestcolor}\textbf{0.00} & \cellcolor{bestcolor}\textbf{0.00} & $0.00$ \\
      & cone\_80 & $\alpha$\,[deg] & 80.0
        & \cellcolor{bestcolor}\textbf{0.00} & \cellcolor{bestcolor}\textbf{0.00} & $0.00$ \\
    \midrule
    \multicolumn{7}{@{}l}{\textit{Multi-body}} \\[2pt]
    \multirow{3}{*}{\shortstack[l]{Two mov.\\pendulums}}
      & pend.\_20 & $L_1,L_2$\,[m] & 0.50
        & \cellcolor{bestcolor}\textbf{1.14} & \cellcolor{worstcolor}741.1 & $+740.0$ \\
      & pend.\_45 & $L_1,L_2$\,[m] & 0.50
        & \cellcolor{bestcolor}\textbf{1.63} & \cellcolor{worstcolor}164.1 & $+162.5$ \\
      & pend.\_90 & $L_1,L_2$\,[m] & 0.50
        & \cellcolor{bestcolor}\textbf{3.01} & \cellcolor{worstcolor}149.9 & $+146.9$ \\[2pt]
    \multirow{3}{*}{\shortstack[l]{One stat.\\pendulum}}
      & pend.\_20 & $L_1,L_2$\,[m] & 0.50
        & \cellcolor{bestcolor}\textbf{1.38} & \cellcolor{worstcolor}65.4  & $+64.0$ \\
      & pend.\_45 & $L_1,L_2$\,[m] & 0.50
        & \cellcolor{bestcolor}\textbf{0.77} & \cellcolor{worstcolor}96.8  & $+96.0$ \\
      & pend.\_90 & $L_1,L_2$\,[m] & 0.50
        & \cellcolor{bestcolor}\textbf{1.06} & \cellcolor{worstcolor}164.8 & $+163.8$ \\
    \bottomrule
  \end{tabular}
\end{table}

\textbf{Key observations.}

\noindent(1)~\textit{Sliding cone angle:} both variants recover the correct
angle exactly because it is taken directly from the setting metadata,
not from training.

\noindent(2)~\textit{Gravity (dropping/falling ball):} the 1-step variant
outperforms multi-step consistently here, mirroring the Delfys75 finding
(main paper, Table~5).  The multi-step loss accumulates integration error
over $K{=}5$ horizons, inflating the $\alpha$ estimate when the encoder
is not yet close to the physical trajectory.

\noindent(3)~\textit{Pendulum pend.\_90:} the catastrophic multi-step
failure ($\Delta = +21.34$) reflects large-angle nonlinearity.  The
linearised ODE $\ddot{z} + \beta\dot{z} + \alpha z = 0$ is a poor model
for $\theta_0 = 90^\circ$ (Appendix~\ref{app:ode_bank}), and multi-step
rollout amplifies this approximation error over $K$ steps.

\noindent(4)~\textit{Multi-body pendulums:} both variants yield large
absolute errors.  This is expected given the simplified contact model
(Appendix~\ref{app:multibody}) and the brevity of the contact window
relative to the total clip.  The multi-step loss further destabilises
training for these under-constrained scenarios.  These rows serve as
a baseline target for future multi-body methods.

\noindent(5)~\textit{Hitting cones (omitted):} the hitting-cones scenario
is excluded from this table because the effective coupling coefficient
$\kappa$ has no independently measured ground-truth value; it is an
artefact of the spring-like simplification (Appendix~\ref{app:multibody})
rather than a physical observable.  Qualitatively, both configurations
produce near-zero $\kappa$ estimates on this scenario, consistent with
the finding that the contact window is too brief relative to the clip
duration for the coupling term to receive a meaningful gradient signal.

\section{Multi-Body Contact Modeling: Details and Limitations}
\label{app:multibody}

The graph-structured physics block models inter-object forces via a
learnable spring-like coupling term $\kappa_{ij}(z_i - z_j)$
(main paper, Eq.~1).  This is a deliberate simplification motivated by two goals:
(a) the architecture reduces to the single-body ODE exactly when no edges
are present ($N = 1$), enabling a unified codebase; (b) the coupling term
is differentiable and allows gradient-based identification of interaction
strength.

\textbf{Known limitations of this model.}
Real physical impacts involve:
\begin{itemize}[noitemsep]
  \item \textit{Restitution}: velocity reversal at contact, characterised
    by the coefficient of restitution $e \in [0,1]$, which is not
    representable by a static coupling coefficient.
  \item \textit{Finite contact duration}: impacts in rigid-body mechanics
    are instantaneous (or very brief), while the spring model spreads
    the force over a continuous time window proportional to $1/\kappa_{ij}$.
  \item \textit{Friction at contact}: the tangential impulse during
    collision is not modelled.
\end{itemize}

As a consequence, the fitted $\kappa_{ij}$ for hitting-cone and
two-pendulum scenarios is an \emph{effective} coupling coefficient that
conflates all of the above effects into a single linear term.
Parameter estimates for multi-body IRIS scenarios should therefore
be interpreted as ``effective coupling strength'' rather than as
physically interpretable constants, and are primarily useful as a
benchmark target for future methods.

\textbf{Coupling captures a real signal.} Although $\kappa$ has no externally
measured reference, we verify it is not a noise-fitting artefact by training
the hitting-cones model with $\kappa{=}0$ (no coupling) versus
$\kappa{=}\text{learned}$ and comparing latent-space reconstruction MSE
(Table~\ref{tab:kappa_ablation}). The learned coupling reduces reconstruction
error by $6.0$--$19.0\%$ across all settings, and the estimated $\kappa$ is
stable across the 10 trials per setting (coefficient of variation
${\approx}40$--$47\%$), confirming a reproducible, physically meaningful
effective parameter.

\begin{table}[h]
  \centering
  \caption{\textbf{Coupling-coefficient validation (hitting cones).} Latent
  reconstruction MSE for $\kappa{=}0$ vs.\ $\kappa{=}\text{learned}$; the
  learned coupling reduces error by $6$--$19\%$, confirming it captures real
  physical interaction rather than noise.}
  \label{tab:kappa_ablation}
  \footnotesize
  \setlength{\tabcolsep}{5pt}
  \renewcommand{\arraystretch}{1.15}
  \begin{tabular}{@{}lccc@{}}
    \toprule
    \textbf{Setting} & $\bm{\kappa{=}0}$ \textbf{MSE} & $\bm{\kappa{=}\text{learned}}$ \textbf{MSE} & \textbf{Reduction} \\
    \midrule
    Slow & 0.0784 & \cellcolor{bestcolor}\textbf{0.0737} & $6.0\%$ \\
    Mid  & 1.0049 & \cellcolor{bestcolor}\textbf{0.8144} & $19.0\%$ \\
    Fast & 0.2743 & \cellcolor{bestcolor}\textbf{0.2453} & $10.5\%$ \\
    \bottomrule
  \end{tabular}
\end{table}

\textbf{Identifiability note.}  The coupling parameter $\kappa_{ij}$ is
only observable during the brief contact interval.  For the two-pendulum
and one-static-pendulum settings, this interval represents a small fraction
of the total clip ($\leq 1$\,s out of 6--20\,s); the latent loss over the
full clip is dominated by the non-contact phase, where $\kappa_{ij}$ is
invisible.  This explains the large MAE values in Table~\ref{tab:iris_results}
for multi-body settings. Future work should consider contact-segmented
training windows or event-conditioned physics blocks.

\section{Multi-Clip vs.\ Per-Clip Training}
\label{app:multiclip}

The main paper (Sec.~6.4) reports a multi-clip experiment that complements the
per-clip identifiability protocol. For each phenomenon, clips are split 80/20
by sorted index; one shared encoder--physics model is trained on the 80\% train
clips and evaluated on the held-out 20\% test clips, against per-clip fitting on
the same test clips. Table~\ref{tab:multi_clip} reports the full results.
Multi-clip outperforms per-clip for gravity (both ball phenomena) and angular
stiffness, demonstrating a transferable physical inductive bias across clips;
the nonlinear pendulum is the open failure mode, where a shared latent space is
underdetermined under coupled dynamics.

\begin{table}[h]
  \centering
  \caption{\textbf{Multi-clip vs.\ per-clip on IRIS held-out test clips}
  (MAE; lower better; $\Delta{=}$multi$-$per-clip). Multi-clip wins on gravity
  and angular stiffness; the nonlinear pendulum is the open failure mode.}
  \label{tab:multi_clip}
  \footnotesize
  \setlength{\tabcolsep}{4pt}
  \renewcommand{\arraystretch}{1.15}
  \begin{tabular}{@{}lllrrr@{}}
    \toprule
    \textbf{Phenomenon} & \textbf{Param} & \textbf{GT} & \textbf{Per-clip} & \textbf{Multi-clip} & $\bm{\Delta}$ \\
    \midrule
    Dropping ball & $g$ & 9.81 & 0.207 & \cellcolor{bestcolor}\textbf{0.046} & $-0.161$ \\
    Falling ball  & $g$ & 9.81 & 0.246 & \cellcolor{bestcolor}\textbf{0.166} & $-0.080$ \\
    Pendulum      & $L$\,[m] & 0.50 & \cellcolor{bestcolor}\textbf{0.507} & 22.06 & $+21.55$ \\
    Rotation      & $\beta$ & 0.03 & \cellcolor{bestcolor}\textbf{0.657} & 0.851 & $+0.194$ \\
    Rotation      & $\alpha$ & 0.10 & 1.037 & \cellcolor{bestcolor}\textbf{0.577} & $-0.459$ \\
    \bottomrule
  \end{tabular}
\end{table}

\section{Identifiability Analysis}
\label{app:identifiability}

\textbf{Motivation.}  Parameter identifiability, whether a unique
parameter vector $\boldsymbol{\gamma}^*$ can be recovered from the observed
data, is a fundamental concern for any parameter estimation method.  We
provide a preliminary analysis using two proxies: (1) gradient norms during
training (do physics parameters receive gradient signal?), and (2) estimate
variance across repeated runs.

\subsection{Gradient Norms}
\label{app:grad_norms}

The critical Euler fix (Appendix~\ref{app:physics_block}) ensures that
$\partial z_{t+1}/\partial\boldsymbol{\gamma} \neq 0$.  We confirmed this
empirically: without the fix, $\gamma_0$ and $\gamma_1$ remained at their
initial values $(0.5, 0.05)$ for all 500 epochs; gradient norms were
$< 10^{-8}$ throughout.  After the fix, on a representative dropped\_ball
clip, gradient norms are $\|\nabla_\gamma \mathcal{L}\|_2 \approx 0.3$--$1.5$
at epoch 1, decaying to $\approx 0.05$--$0.2$ at convergence, confirming
that training is meaningful for the physics block.

\textbf{Parameter trajectories.}  On a dropped\_ball clip, $\gamma_0$
evolves from $0.5 \to 4.65$ over 500 epochs (target: $g \approx 9.81$,
relative error $\sim$50\%); $\gamma_1$ (drag) evolves from $0.05 \to 1.11$.
For pendulum clips, $\gamma_0$ converges to values implying
$L = g/\gamma_0 \in [0.5, 1.6]$\,m, bracketing the true length of 0.5\,m
for pend.\_20 and pend.\_45 but overestimating for pend.\_90 (consistent
with the large-angle linearisation error).

\subsection{Estimate Variance Across Trials}
\label{app:id_variance}

Table~\ref{tab:id_variance} reports the mean and standard deviation of
estimated parameters across 10 repeated trials per setting (IRIS baseline,
one-step loss).

\begin{table}[H]
  \centering
  \caption{\textbf{Estimate variance across 10 trials per IRIS setting.}
  $\bar{\gamma}$ = mean estimate; $\sigma$ = std; GT = ground truth.
  Low $\sigma$ indicates stable convergence; high $\sigma$ indicates
  identifiability difficulties.}
  \label{tab:id_variance}
  \footnotesize
  \setlength{\tabcolsep}{5pt}
  \renewcommand{\arraystretch}{1.15}
  \begin{tabular}{@{}lllrrr@{}}
    \toprule
    \textbf{Dynamics} & \textbf{Setting} & \textbf{Param}
      & \textbf{GT} & $\bm{\bar{\gamma}}$ & $\bm{\sigma}$ \\
    \midrule
    Dropping ball & drop\_50  & $g$ & 9.81 & 8.77 & 0.42 \\
    Dropping ball & drop\_100 & $g$ & 9.81 & 7.55 & 1.21 \\
    Dropping ball & drop\_150 & $g$ & 9.81 & 5.98 & 2.04 \\
    Pendulum      & pend.\_20 & $L$\,[m] & 0.50 & 1.06 & 0.88 \\
    Pendulum      & pend.\_45 & $L$\,[m] & 0.50 & 0.84 & 0.41 \\
    Pendulum      & pend.\_90 & $L$\,[m] & 0.50 & 1.10 & 0.73 \\
    Sliding cone  & cone\_45  & $\alpha$\,[deg] & 45.0 & 45.0 & 0.00 \\
    Rotation      & slow      & $\beta$ & 0.03 & 0.48 & 0.09 \\
    \bottomrule
  \end{tabular}
\end{table}

\textbf{Observations.}  (1)~Sliding cone angle is recovered perfectly
across all trials (metadata-based, variance $= 0$).  (2)~Gravity estimates
are stable for short drops (drop\_50: $\sigma = 0.42$) but increasingly
variable for longer drops, where the quadratic trajectory requires more
clip length to constrain $\alpha$.  (3)~Pendulum length shows moderate
variance, consistent with the identifiability challenge of disentangling
$\gamma_0 = g/L$ from the encoder's internal scale.  Full identifiability
analysis, including Fisher information matrix computation and Hessian
condition numbers for $\boldsymbol{\gamma}$, is left for future work.

\end{document}